\documentclass[acmsmall,nonacm]{acmart}
\usepackage{fontawesome}
\usepackage{tabularx}
\usepackage{multirow}
\usepackage{booktabs}
\usepackage{ragged2e}
\usepackage{graphicx}
\usepackage{wrapfig} 
\AtBeginDocument{%
  }

\setcopyright{acmlicensed}
\copyrightyear{2018}
\acmYear{2018}
\acmDOI{XXXXXXX.XXXXXXX}

\acmJournal{JACM}
\acmVolume{37}
\acmNumber{4}
\acmArticle{111}
\acmMonth{8}

\begin{document}

\title{Hyperspectral Anomaly Detection Methods: A Survey and Comparative Study}

\author{Aayushma Pant}
\author{Arbind Agrahari Baniya}
\author{Tsz-Kwan Lee}
\author{Sunil Aryal}
\email{{a.pant,a.agraharibaniya,glory.lee,sunil.aryal}@deakin.edu.au}

\affiliation{%
  \institution{\\School of Information Technology, Deakin University}
  \city{Melbourne}
  \state{VIC}
  \country{Australia}
}

\renewcommand{\shortauthors}{A.Pant et al.}

\begin{abstract}
 
Hyperspectral images are high-dimensional datasets consisting of hundreds of contiguous spectral bands, enabling detailed material and surface analysis. Hyperspectral anomaly detection (HAD) refers to the technique of identifying and locating anomalous targets in such data without prior information about a hyperspectral scene or target spectrum. This technology has seen rapid advancements in recent years, with applications in agriculture, defence, military surveillance, and environmental monitoring. Despite this significant progress, existing HAD methods continue to face challenges such as high computational complexity, sensitivity to noise, and limited generalisation across diverse datasets. This study presents a comprehensive comparison of various HAD techniques, categorising them into statistical models, representation-based methods, classical machine learning approaches, and deep learning models. We evaluated these methods across 17 benchmarking datasets using different performance metrics, such as ROC, AUC, and separability map to analyse detection accuracy, computational efficiency, their strengths, limitations, and directions for future research. Our findings highlight that deep learning models achieved the highest detection accuracy, while statistical models demonstrated exceptional speed across all datasets. This survey aims to provide valuable insights for researchers and practitioners working to advance the field of hyperspectral anomaly detection methods. 
 
\end{abstract}

\ccsdesc[100]{Computing methodologies~Scene anomaly detection}

\keywords{Hyperspectral Imaging, Artificial Intelligence, Anomaly Detection}


\maketitle

\section{Introduction}
\subsection{Background}
Hyperspectral imaging (HSI) is an advanced remote sensing technology that captures narrow spectral bands across a broad range of the electromagnetic spectrum, providing detailed spectral information about materials and objects. Unlike conventional visible images that rely on a limited number of spectral bands (e.g., RGB channels) or multispectral images with a restricted spectral resolution, HSI provides rich spectral signatures for each pixel, making it a powerful tool for material identification, environmental monitoring, and anomaly detection. Hyperspectral images are represented as a 3D data cube, where two dimensions $(x,y)$ correspond to spatial information and the third dimension $(z)$ corresponds to spectral bands ~\cite{Su_2022}. Hyperspectral images rely on the fact that every material possesses a unique spectral signature, which refers to the ratio of reflected and incident electromagnetic radiation variation as a function of wavelength. Analysis of such hyperspectral images includes spectral unmixing to separate contributions of different objects in given pixels, hyperspectral object classification that classifies each pixel into a particular class, and anomaly detection to detect pixels with unusual spectral characteristics \cite{LetianDENG_11}. Anomalies in hyperspectral images refer to objects or regions of interest that exhibit unusual spectral patterns, which deviate from the background or dominant signatures typically present in the scene, referred to as the norm \cite{Xu_2022}. These 'normal' spectral patterns are often defined by the statistical properties or abundance of the majority class, representing the expected spectral response of common patterns (intensities, compositions, reflectance properties, etc.) in the image. The anomalies typically cover only a negligible area of the image. In real-world applications, anomalies can take various forms depending on the industry and the expected background of the scene. For example, in defence and aerospace, the typical background might be vast landscape consisting of ocean, desert or forest. In such contexts, anomalies would include man-made objects like  aircraft, space debris, ships, vehicles, as well as distinct features like rooftops, because their spectral properties are markedly different from the dominant natural or urban background. These anomaly detection applications can be used in search and rescue mechanisms, mineral reconnaissance, defence surveillance, and hazard detection \cite{10847782}. In the agriculture sector, where a 'normal' scene consists of healthy crops or specific soil types, anomalies can indicate crop disease or soil contamination due to their altered spectral responses. Similarly, in food safety, hyperspectral imaging can detect chemical adulteration \cite{RAM2024109037}, such as excessive urea in milk, where the adulterant's spectral signature deviates from that of pure milk.
\par This paper focuses on HAD in airborne remote sensing and satellite-based datasets, such as AVIRIS (Airborne Visible Infrared Imaging Spectrometer), HYDICE (Hyperspectral Digital Imagery Collection Experiment), and ABU (Airport Beach and Urban Dataset), which are commonly used for detecting airborne anomalies against natural backgrounds for remote sensing applications.
\begin{wrapfigure}[13]{r}{0.5\textwidth} 
    \centering
    \includegraphics[width=\linewidth]{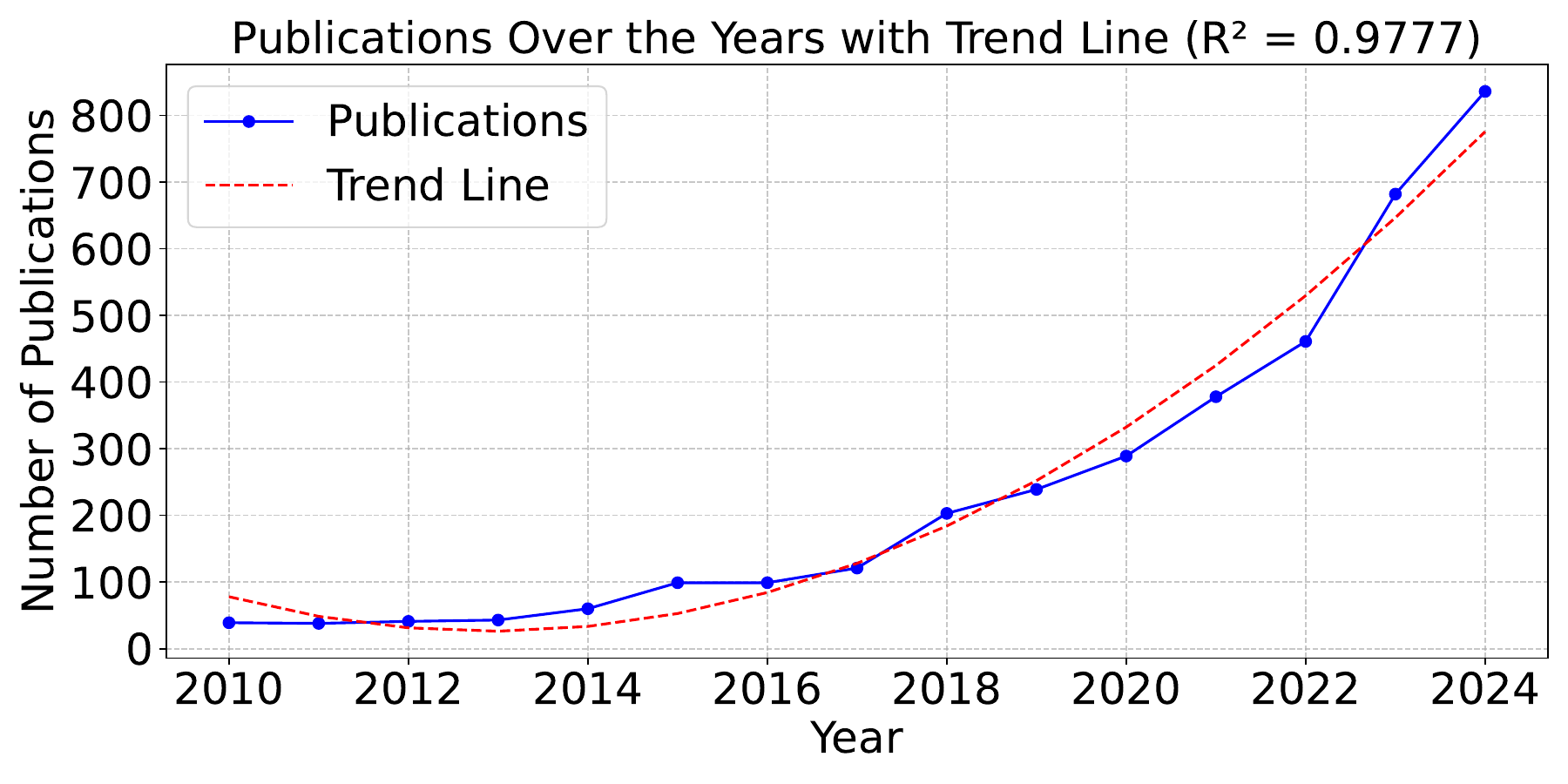} %
    \caption{Exponentially growing trend in Hyperspectral Anomaly Detection from 2010 to 2024}
    \label{fig:fig1}
\end{wrapfigure}
With the growing body of research in HAD, the field has experienced remarkable progress in recent years. As shown in Figure \ref{fig:fig1}, the number of articles published on HAD has increased exponentially from approximately 80 in 2010 to over 900 in 2024 based on data from Scopus and Google Scholar. This upward trend is supported by a strong coefficient of determination ($ \mathbf{R}^ 2  \approx 0.97 $ ), indicating a consistent and significant rise in research interest. This growth reflects the continuous emergence of novel HAD algorithms that address key challenges such as complex background scenarios in images, limited data availability, computational inefficiency, and the difficulty of accurately detecting anomalies.
\par Despite this significant progress and the increasing use of advanced deep learning algorithms like diffusion models \cite{ma_bsdm_2023,10858750}, attention-driven models \cite{10506667,lian_gt-had_2024,10073635}, and state space models (Mamba) \cite{He_2024, Huang_2024} in HAD, existing review articles \cite{5546306, su_hyperspectral_2021, xu_hyperspectral_2022, rs14091973, RazaShah31122022} often present fragmented overviews. Many surveys primarily focus on algorithm descriptions without adequately addressing how hyperspectral features are learned and utilised, especially in mitigating issues like data noise and scarcity.  Moreover, they frequently fail to provide a comprehensive discussion on the evolving role of deep learning, model scalability, and real-world applicability, or offer only high-level summaries lacking in-depth analysis. Runtime efficiency and computational complexity, which are critical factors for real-world deployment, are also often overlooked.
\begin{table*}[ht]
\caption{Comparison of this article with other reviews on HAD studies}
\Description{Seven‐column comparison table showing how fully each review considers various aspects, using circle icons filled 0\%, 50\%, or 100\%.}
\label{table:relatedwork}
\footnotesize
\begin{tabularx}{\linewidth}{c> {\centering\arraybackslash}X > {\centering\arraybackslash}X >{\centering\arraybackslash}X >{\centering\arraybackslash}X >{\centering\arraybackslash}X >{\centering\arraybackslash}X}

\toprule
\textbf{Reference} & \textbf{Year} & \textbf{Noise Removal} & \textbf{Deep Learning } & \textbf{Dataset Mentioned } & \textbf{Time Complexity} & \textbf{Performance Comparison} \\
\midrule
\cite{5546306} & 2010 & \textit{\faCircleO} & \textit{\faCircleO} & \textit{\faCircleO} & \textit{\faCircleO} & \textit{\faCircleO} \\
\cite{su_hyperspectral_2021} & 2022 & \textit{\faCircleO} & \textit{\faAdjust} & \textit{\faCircle} & \textit{\faAdjust} & \textit{\faCircleO} \\
\cite{xu_hyperspectral_2022} & 2022 & \textit{\faCircleO} & \textit{\faAdjust} & \textit{\faCircleO} & \textit{\faCircleO} & \textit{\faCircleO} \\
\cite{rs14091973} & 2022 & \textit{\faCircleO} & \textit{\faAdjust} & \textit{\faCircle} & \textit{\faCircleO} & \textit{\faAdjust} \\
\cite{RazaShah31122022} & 2022 & \textit{\faCircleO} &\textit{\faCircleO} & \textit{\faCircle} & \textit{\faCircleO} & \textit{\faAdjust} \\
\cite{rs16203879} & 2024 & \textit{\faCircleO} & \textit{\faCircleO} & \textit{\faCircleO} & \textit{\faCircleO} & \textit{\faCircleO}\\
Ours & 2025 & \textit{\faCircle} & \textit{\faCircle} & \textit{\faCircle} & \textit{\faCircle} & \textit{\faCircle} \\
\bottomrule
\end{tabularx}
\footnotesize
\raggedright
\textit{Legend:} \faCircle~Fully considered, \faAdjust~Partially considered, \faCircleO~Sparsely or Not considered.
\end{table*}

\par To bridge these identified gaps and provide a robust guide to the evolving research landscape of HAD, this paper offers a review with several key contributions:

\begin{itemize}
    \item A structured and detailed categorisation of HAD algorithms into four major groups: deep learning-based approaches (further grouped into six broad subcategories based on their architectural characteristics), statistical methods, representation-based models, and classical machine learning frameworks.
    \item In-depth analysis of feature learning mechanisms and a discussion on deep learning algorithms, including their scalability and real-world applicability.
    \item Extensive coverage of over 17 widely used datasets, presented with a comprehensive table detailing each dataset’s characteristics, including anomaly type, citations, data source, and format.
    \item A comparative evaluation across algorithms from each category, utilising standard metrics such as ROC curves, AUC scores, and box-and-whisker plots to ensure a fair and consistent analysis.
    \item Detailed coverage of critical real-world factors such as noise removal techniques, runtime efficiency, which are often overlooked in existing research.
    \item Practical recommendations, discussions on persistent challenges, and suggestions for future research directions.
\end{itemize}
Table \ref{table:relatedwork} further summarises how this review addresses these limitations and distinguishes itself from existing survey articles by providing more extensive and in-depth coverage across crucial aspects of HAD research.

\subsection{Organisation}
This review is organised to provide a logical progression through the field of HAD. We begin in Section~\ref{sec:taxonomy} by establishing the framework for our discussion, outlining the review methodology, and introducing a new taxonomy for categorising HAD approaches.  Section~\ref{sec:methods} then explores the fundamental techniques, covering a spectrum from statistical methods to modern deep learning approaches. To reflect the evolving nature of HAD, Section~\ref{sec:trends} illuminates recent developments, including novel input variations, advanced feature engineering techniques for hyperspectral images, and improved noise handling algorithms. A significant contribution of this work is presented in Section~\ref{sec:analysis}, which provides an in-depth evaluation of 17 publicly available HAD datasets, featuring a comprehensive comparative analysis of algorithms, along with insights into their computational complexity and runtime. Sections~\ref{sec:challenges} and~\ref{sec:future_enhance} address current limitations in HAD research, propose actionable recommendations, and suggest directions for future research. The paper concludes in Section~\ref{sec:conclusion} with a concise summary of the key insights derived from this review.

\section {Research Methodology and Taxonomy}~\label{sec:taxonomy}
This review employs a \textbf{narrative synthesis approach}, chosen for its flexibility in integrating diverse research findings and providing a holistic understanding of the rapidly evolving field of HAD. Our primary objective is to not only summarise existing HAD methods but also to provide a structured and critical evaluation of their underlying principles, strengths, limitations, and practical applicability within real-world scenarios.
\par We performed a systematic literature search across major academic databases, primarily Google Scholar and Scopus. We utilised their advanced search functionalities to apply a broad spectrum of keyword combinations, including: \texttt{"hyperspectral anomaly detection" AND ("remote sensing" OR "satellite" OR "aerial imaging" OR "drone" OR "UAV" OR "earth observation" OR "geospatial analysis" OR "environmental monitoring" OR "urban planning" OR "land cover" OR "agriculture" OR "disaster management" OR "military surveillance" OR "forest monitoring")}. This initial search was complemented by a forward and backward snowballing technique \cite{WOHLIN2022106908}, where reference lists of highly cited and foundational HAD articles were meticulously reviewed to identify additional relevant studies, until saturation was reached. To ensure the review reflects the most current and impactful advances, we primarily focused on publications from 2015 to early 2025, a period characterised by significant growth and diversification in HAD research, particularly with the rise of deep learning. 
Furthermore, to identify widely used benchmark datasets, we used queries such as \texttt{"hyperspectral anomaly detection datasets"} and \texttt{"AVIRIS datasets"}.
\par Our comparative analysis of HAD algorithms as shown in Figure~\ref{fig:taxonomy}, categorises them according to a novel methodological taxonomy we propose: (i) statistical methods  (ii) representation based methods (iii) classical machine learning models and (iv) deep learning models. These categories reflect the distinct principles and computational frameworks that have shaped the evolution of HAD research. For each category, we examine algorithms based on several key criteria: quantitative performance metrics (including ROC curves, AUC scores, background separability, and false alarm rate), robustness to noise, input types and computational efficiency (including both complexity and practical runtime). Our synthesis integrates both these quantitative and qualitative evaluations, such as architectural complexity, data dependency, and suitability for diverse application contexts, to offer a holistic view of the HAD research landscape. 

\section{Algorithms for Hyperspectral Anomaly Detection}~\label{sec:methods}
\begin{figure}[htbp!]
    \centering
    \includegraphics[width=1\linewidth]{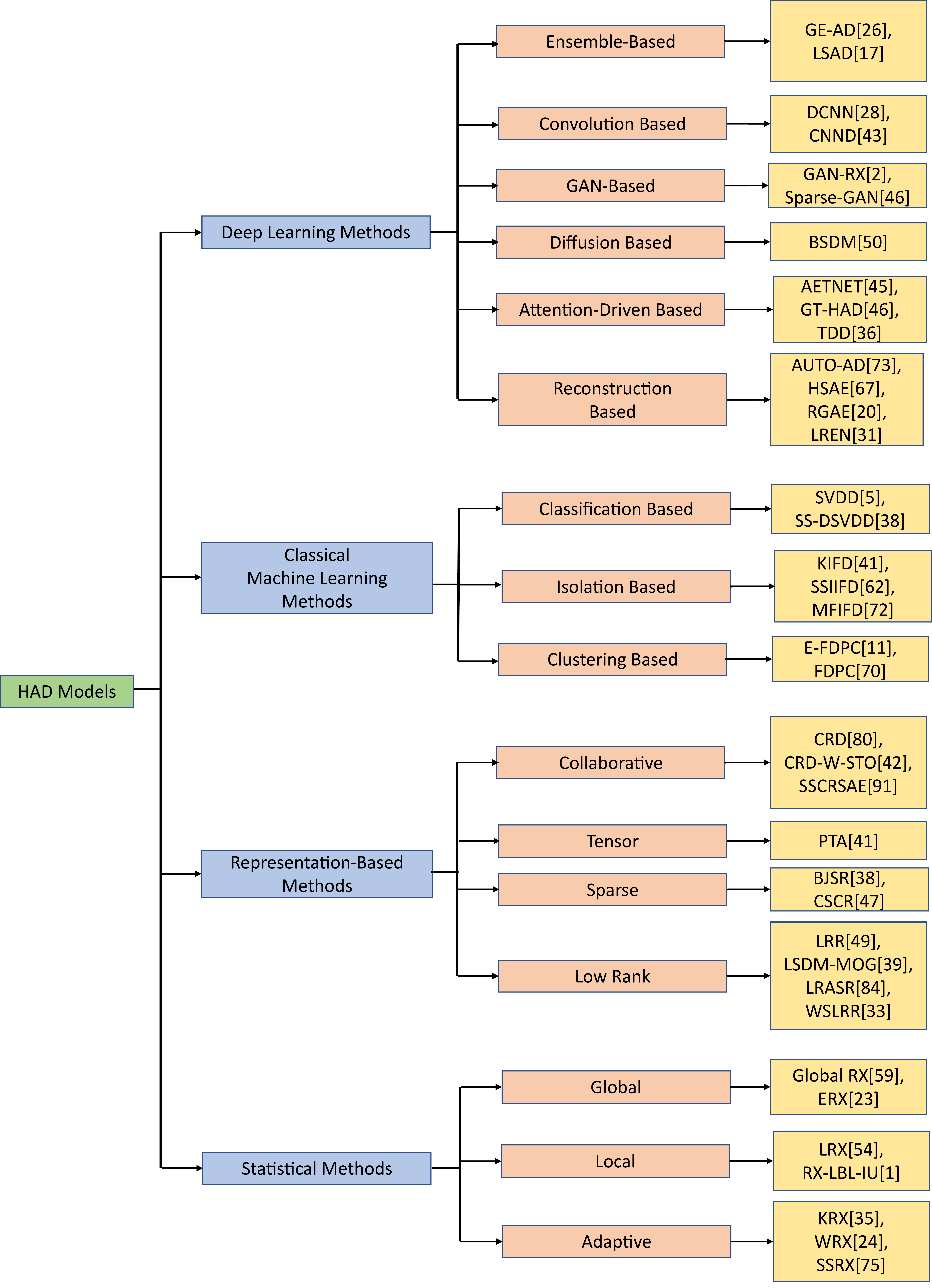}
    \caption{Our Proposed Taxonomy for Hyperspectral Anomaly Detection Algorithms. This taxonomy provides a structured overview, classifying methods based on their methodologies, offering a unique perspective for this survey.  }
    \label{fig:taxonomy}
\end{figure}

As highlighted in Section~\ref{sec:taxonomy}, algorithms for HAD are systematically categorised into four principal groups: statistical methods, representation-based methods, classical machine learning models, and deep learning based models, as visually summarised in Figure~\ref{fig:taxonomy}. Statistical methods utilise statistical properties, such as mean vectors and covariance matrices, to model the background distribution and  detect anomalies as pixels that deviate significantly from this model and are further classified into global, local, and adaptive approaches. Representation-based methods, on the other hand, construct a background dictionary or basis to represent normal pixels, and identify anomalies as samples that cannot be accurately represented by this dictionary, with subcategories including collaborative, sparse, tensor-based, and low-rank representation models. Classical machine learning models apply established machine learning techniques, focusing on pattern recognition,  clustering principles to differentiate anomalies from background, such as Support Vector Machines (SVM) and Isolation Forests. SVMs use hyperplane separation in high-dimensional space, while Isolation Forests rely on randomly constructed trees (iTrees) to isolate anomalies based on recursive feature splits. Finally, Deep learning-based methods utilise multi-layered neural networks to automatically extract hierarchical spatial and spectral features for robust anomaly discrimination. These approaches are further sub-categorised into six prominent paradigms: convolutional neural network (CNN)-based models, generative adversarial network (GAN)-based models, ensemble-based models, attention-based models, diffusion-based models, and reconstruction-based models.

\subsection{Statistical Methods}
Statistical models form the foundational layer of HAD, utilising the statistical properties of hyperspectral data to distinguish between background and anomalous pixels. These methods fundamentally operate by characterising the normal background using probabilistic distributions and statistical measures (such as mean vectors and covariance matrices). Anomalies are then identified as pixels that significantly deviate from this learned background distribution. These methods are often known for their simple, strong and fast computation. These methods are broadly categorised based on the scope of their background statistics estimation:

\subsubsection{Global Statistical Methods}
Global statistical methods estimate background statistics using the entire hyperspectral image. This approach assumes that the background within the scene is relatively homogeneous following a multivariate Gaussian distribution. While offering computational efficiency due to a single, scene-wide calculation, their performance typically degrades in complex, heterogeneous environments where the underlying multivariate Gaussian assumption is often violated, leading to increased false alarm rates. The prominent global statistical techniques include:

\paragraph{Global Reed-Xiaoli (GRX) Detector}

The RX algorithm \cite{60107}  stands as a fundamental benchmark for HAD. It assumes that background pixels follow a multivariate Gaussian distribution, enabling statistical modelling of the background. The RX detector calculates the probability density function by calculating the mean vector ($\boldsymbol{\mu}$) and covariance matrix ($\boldsymbol{\Sigma}$) from all pixels and estimates the Mahalanobis distance between test pixels ($x$ ) and the background to identify anomalies.

\begin{equation}
D_{\text{RX}}(x) = (x - \boldsymbol{\mu})^T \boldsymbol{\Sigma}^{-1} (x - \boldsymbol{\mu}),
\end{equation}


Despite the effectiveness in simple, homogeneous environments, the RX detector does not model higher-order dependencies among spectral bands, leading to high false positives in cluttered environments.

\paragraph{ERX (Exponential RX Algorithm)}
This algorithm \cite{10847782}  is designed for improved computational efficiency and real-time anomaly detection that operates in a streaming fashion using hyperspectral line-scan cameras. It utilises sparse random projections for dimensionality reduction and employs exponentially weighted moving averages to continuously update the global background mean and covariance statistics pixel-by-pixel. While significantly accelerating processing for large datasets, its reliance on pixel-by-pixel inference may limit its robustness in detecting anomalies in complex backgrounds by potentially overlooking broader spatial context.

\subsubsection{Local Statistical Methods}
Local statistical methods overcome the limitations of global approaches by estimating background statistics from a specific, localised area rather than the entire image. This adaptability allows them to account for localised background heterogeneity, making them more robust in complex scenes and where anomalies are small. A common challenge in these methods involves precisely defining the optimal local neighbourhood, often through the use of window structures, to ensure the background model accurately reflects normal local variations without being contaminated by the anomaly itself.

\paragraph{Local RX (LRX)}
LRX \cite{6851148} employs a double-window (inner and outer) structure that slides across the image. For a central pixel, its background statistics (mean and covariance matrix) are computed exclusively from the outer window. The inner window acts as a guard region, preventing potential anomalies from contaminating the local background estimation. This design significantly improves the detection of small and sub-pixel anomalies in heterogeneous environments.
For each test pixel $\mathbf{x}$, a local window of size $(k \times k)$ is centred at $(x)$ to facilitate the detection of small and sub-pixel anomalies \cite{Raza_Shah_2022}. The general form of covariance matrix calculation often seen in such methods can be represented as:

\begin{equation} 
   \boldsymbol{\Sigma} =
    \begin{cases} 
        \boldsymbol{\Sigma} = \frac{X^T X}{n}, & \text{for the RX algorithm,} \\
        \boldsymbol{\Sigma}_{\kappa}(\mathbf{x}) = \frac{X_{\kappa}^T X_{\kappa}}{n_{\kappa}}, & \text{for the LRX algorithm}.
    \end{cases}
\end{equation}

Here $X$ represents the hyperspectral image with $X^T = [\mathbf{x}_1,\mathbf{x}_2, \dots, \mathbf{x}_n]$, denoting the matrix of $n = l \times s$ pixel vectors. The subscript $\kappa$ denotes the pixels included within the local sliding window for the LRX algorithm.
\par Different improvements in local anomaly detection algorithms have emerged over time. For instance, RX-LBL-IU (line-by-line RX with inverse update) \cite{10.1007/s11554-011-0205-x} uses a "sliding" double-window approach around each pixel, but critically, it applies the Woodbury matrix identity to recursively update the window inverse covariance, rather than recalculating it entirely for each pixel's window. Another method, Dark HORSE RX \cite{10.1007/s11554-011-0205-x}, utilises exponentially rolling mean and covariance for efficient pixel-wise updates.

Despite its enhanced adaptability, LRX methods face a major challenge in calculating the window statistics. Incorrect sizes can either suppress small anomalies if the inner window is too large or include background heterogeneity if the outer window is too small, often requiring manual tuning for optimal performance. Additionally, small window sizes can lead to matrix singularities due to insufficient samples for covariance estimation, while large window sizes slow the computation time. 

\subsubsection{Adaptive Statistical Methods}
Adaptive statistical methods extend the basic assumptions of traditional RX-based detectors by incorporating more sophisticated mathematical and statistical concepts. These approaches aim to address fundamental limitations such as non-Gaussianity, non-linearity in spectral relationships, or noise sensitivity, often by transforming the data or employing more complex probabilistic models.

\paragraph{Kernel RX (KRX)}

KRX\cite{1386510} extends the linear RX detector by mapping the input hyperspectral data into a higher-dimensional feature space using a nonlinear kernel function (e.g., Radial Basis Function, Polynomial). In this transformed space, complex nonlinear spectral relationships might become linearly separable, allowing the RX detection principle to be applied more effectively. However, its performance is sensitive to the choice and parameter tuning of the kernel function, requiring more matrix operations that slow down the calculations.

\paragraph{Weighted RX (WRX)}
WRX\cite{6782328} enhances anomaly detection performance of RX method by assigning different weights to background pixels during the covariance matrix estimation. Instead of treating all background pixels equally, WRX assigns a higher weight to the pixels closer to the background than the pixels far away from the background. This weighting mechanism  reduces the influence of anomalies or noise in the background estimation, leading to fewer false positives in cluttered environments.
\paragraph{Subspace RX (SSRX)}
SSRX\cite{6002215} addresses the high dimensionality and inherent structure of hyperspectral data by first projecting the data into a lower-dimensional subspace. This dimensionality reduction is  achieved using techniques like adaptive subspace decomposition\cite{899445,9709262}, Principal Component Analysis \cite{10.1117/12.518835}, where the dominant components are assumed to  capture background variations. Anomalies are then detected by applying an RX-like detector within this derived subspace, or by analysing the reconstruction error from the subspace, assuming anomalies cannot be well-represented by the background subspace.

Despite their effectiveness and increased flexibility over global and local methods, such adaptive statistical algorithms often introduce additional parameters requiring careful tuning and can incur higher computational costs due to their increased model complexity.

\subsection{Representation-Based Methods}  

Representation-based methods for HAD operate on the theory that normal background pixels can be effectively characterised or reconstructed using a learned dictionary or basis, while anomalous pixels cannot. The degree of anomaly is subsequently determined by the reconstruction residuals. A general reconstruction model is expressed as:
\[{x}\approx{D}\mathbf{\alpha}\] where $x$ represents a pixel vector,  $D$ is the learned dictionary, and  $\alpha$ is the coefficient vector (often with regularisation constraints) used for reconstruction.
\par The core idea behind these methods is that background structures exhibit inherent properties, such as low-rank or sparsity, when represented in a suitable basis.  Hyperspectral data, often represented as high-dimensional matrices or tensors, are utilised to derive lower-dimensional approximations. These low-dimensional structures then function as background dictionaries, with anomalies being determined by their sparse representation residuals \cite{Cui2014AnomalyDI,Xu2016AnomalyDI}. Such approaches offer greater flexibility in handling nonlinearity and can capture both global and local image structures through decomposition techniques applied at the superpixel or local region level. The most widely utilised representation-based methods in HAD include low-rank representation, sparse representation, collaborative representation, and tensor representation:

\subsubsection{Sparse Representation}
Sparse representation-based methods \cite{Ling2019ACS} assume that background pixels can be expressed as a sparse linear combination of elements from a learned or pre-defined over complete dictionary, while anomaly pixels, are rare and different for each other are assumed to be non-sparse or to have high representation residuals with respect to this background dictionary. The degree of anomaly is then directly determined by the magnitude of this residual representation from which the anomaly detection is conducted in a hyperspectral image.

\subsubsection{Collaborative Representation} 
While SR aims for the best representation using a minimal number of dictionary elements, CR methods \cite{9826842,6876207,9329029} approach the problem differently. They utilise all atoms in a dictionary to estimate the collaboration between them, modelling each background pixel as a dictionary element derived from its surrounding regions using dual-framed windows. This approach operates under the assumption that background pixels can be reconstructed using a linear combination of their spatial neighbours (which often share similar spectral characteristics and covariance structures), whereas anomalous pixels cannot be similarly represented.

\subsubsection {Low-Rank Representation (LRR)}
Low-Rank Representation (LRR) methods(\cite{10.5555/3104322.3104407,9444588,7322257,9011733} attempt to find the lowest-rank representation of all pixels within a hyperspectral image. This approach models the global structure of the background. In hyperspectral imagery, high spatial and spectral correlation often introduces global low-rank characteristics. For anomaly detection, the fundamental assumption is that the background exhibits a low-rank property due to its inherent spatial homogeneity and spectral correlation, while anomalies, being in small proportion, are sparse. Anomalies are detected by analysing the residual error after subtracting the recovered low-rank background from the original image.

 \subsubsection{Tensor Decomposition} Some advanced representation-based approaches utilise tensor-based models to exploit the inherent multi-dimensional structure (spatial, spectral) of hyperspectral data. Methods like the Prior-Based Tensor Approximation (PTA) \cite{9288702,8625394}decompose hyperspectral images into distinct background and anomaly tensors. This typically involves enforcing low-rank and piecewise-smooth priors for the background representation, while anomalies are modelled using sparse priors at the spatial group level, allowing for more nuanced anomaly separation within the multi-dimensional data space.

These various representation techniques can also be combined to further enhance detection accuracy and robustness. For instance, \cite{SU2020195} integrated low-rank and collaborative representations by utilizing a background dictionary alongside a distance-weighted matrix for improved anomaly isolation. Similarly, \cite{Feng2021AHA} combined low-rank and sparse decomposition with density peak-guided collaborative representation, demonstrating the benefits of using both sparse and low-rank components in a modelling hybrid framework. Such hybrid approaches have consistently shown promising results in isolating anomalies from complex hyperspectral backgrounds.

\par Despite their significant advantages in modeling background structure, representation-based methods have notable limitations. They frequently demand computationally expensive matrix decompositions and iterative dictionary learning processes, which can make them inefficient for large-scale hyperspectral datasets. Furthermore, their effectiveness is heavily dependent on parameter selection, such as dictionary size, sparsity regularisation, or rank approximation, which can be challenging without extensive prior domain knowledge. These methods also implicitly assume that background pixels follow certain low-rank or sparse properties, an assumption that may not consistently hold in highly complex or extremely heterogeneous real-world scenarios. Additionally, managing the high-dimensionality of hyperspectral data remains a challenge, particularly when dealing with irrelevant attributes or noise, which can degrade representation quality.

\subsection{Classical Machine Learning Models}

While representation-based methods rely on predefined mathematical structures for background modelling, classical machine learning techniques offer an alternative by learning patterns directly from data distributions, without requiring explicit prior assumptions about the background's statistical properties. These methods use statistical learning and optimisation techniques to classify or identify anomalies within hyperspectral imagery. Classical machine learning approaches for hyperspectral anomaly detection include a range of paradigms, including classification-based\cite{1661816,9104925}, isolation-based\cite{4781136}, and clustering-based methods\cite{6999481}.

\subsubsection{Classification-based Anomaly Detection}
Classification-based anomaly detection methods train the classifier on the known background data or with labelled anomalies to learn the characteristics of normal samples. Anomalies are then identified as instances that either fall outside the learned normal class boundary or are explicitly classified as abnormal by the trained model. Various classification techniques are employed in HAD, including Support Vector Machines.
\paragraph{Support Vector Machine}
SVM-based methods in HAD adopt a one-class classifier, specifically the Support Vector Data Description (SVDD) \cite{SeifiMajdar2017APS, 1661816}. SVDD works by estimating the smallest hypersphere that encloses the majority of the background samples in a high-dimensional feature space. Pixels lying outside this hypersphere are then considered anomalies, with their distance from the sphere indicating their anomaly level. This approach is  advantageous as it removes the assumption that the background must be Gaussian and homogeneous, and it avoids strong prior assumptions about data distribution. As a supervised machine learning method, SVM also increases robustness to high-dimensional data and does not typically require a large number of training samples. For instance, ss-DSVDD \cite{9684450} utilises an autoencoder alongside an SVM model for enhanced feature learning and anomaly detection. However, a significant challenge in SVM-based methods is the optimisation of kernel parameters and the complex computational process, both of which directly impact the accuracy of the classification results \cite{Chi2007ClassificationOH, Xiang2020HyperspectralAD}.   

\subsubsection{Isolation-based Anomaly Detection}
Isolation-based methods operate on the principle that anomalies are few and different, making them easier to isolate from the majority of normal data points than normal instances are from each other. These techniques focus on separating outliers through recursive partitioning, rather than relying heavily on density estimation or distance metrics like other paradigms.
\par To overcome limitations related to scalability and low-dimensionality in earlier anomaly detection methods, the Isolation Forest (IForest) \cite{4781136} was introduced. IForest explicitly isolates anomalies by recursively partitioning the sample space using binary splits until each instance is either isolated or a maximum path length (tree height) is reached. Compared to many other classical anomaly detection techniques, IForest is highly scalable, efficiently handles large datasets and high-dimensional problems (even with irrelevant attributes), and has linear time complexity, low memory requirements, robustness to noise, and an unsupervised learning capability \cite{cao2024anomalydetectionbasedisolation}. The core principle of Isolation Forest is that anomalies, being rare and distinct, are more susceptible to isolation and tend to have shorter path lengths in the tree structure. Building on this principle, several variations have been developed in hyperspectral data.


 \par Kernel Isolation Forest-Based Detector:  KIFD\cite{8833502} maps hyperspectral data into a kernel space, using both local and global information. By projecting input data into a higher-dimensional feature space, it enhances the computational capability of linear models. The algorithm estimates kernel principal components (KPCA) and uses the most significant KPCA to generate an Isolation Forest for anomaly detection.

\par Multi-Feature Isolation Forest \cite{9040873}: This approach integrates multiple feature types—including spectral, Gabor, extended morphological profile (EMP), and extended multi-attribute profile (EMAP) features. It constructs an Isolation Forest for each feature using a subsampling strategy and combines them to exploit both spectral and spatial information in hyperspectral images.

\par Spatial-Spectral Improved Isolation Forest Detector (SSIIFD) \cite{9521674}: This is an enhanced version that employs the Relative Mass Isolation Forest (ReMassiForest) detector to enhance spatial anomaly detection within hyperspectral images.

Beyond these adaptations, the success of Isolation Forest has led to broader methodological advancements. One such method is iNNE (isolation using Nearest Neighbour Ensemble) \cite{https://doi.org/10.1111/coin.12156}, which isolates each instance from the rest within a subsample defined by its distance to its nearest neighbour (NN) at the centre. This method was developed to address a key limitation of the Isolation Forest, which is its insensitivity to local anomalies and its inability to detect anomalies in low-relevant dimensions, as random subsets may lack sufficient relevant features for effective isolation. Building on this approach, \cite{10553340} further developed iNNE, enhancing spatial feature extraction by applying a Gabor filter to a PCA-projected subspace and using iNNE to derive the spectral anomaly score.
\subsubsection{Clustering-based Anomaly Detection}
Clustering-based methods are well-suited for HAD due to the inherent nature of hyperspectral data. These approaches operate under the assumption that normal data points will form dense, well-defined clusters, while anomalies, being rare and distinct, will either not belong to any major cluster, reside far from cluster centroids, or form very small, sparse clusters themselves. Anomalies are thus identified by analysing density, or distance relative to the discovered clusters.
\par Traditionally, centroid-based clustering algorithms like K-means partition data, assigning each pixel to a class. Outliers are typically considered points deviating significantly from their assigned cluster's centre. However, more recently, density-based clustering has gained popularity in HAD \cite{TU2020144, chen2017hyperspectral, tang2015enhanced}. This shift is driven by the computational expense of calculating global density for every pixel across the entire image; instead, density is often computed within local windows.
\par In density-based approaches, the calculated density score serves as the abnormal level of a test pixel. A lower density score indicates a higher likelihood that a pixel belongs to an abnormal target. This methodology is effective for detecting small targets, though its accuracy for larger targets remains dependent on the appropriate setting of parameters.
 
\subsection{Deep Learning Methods}

While methods like representation-based techniques (e.g., low-rank, sparse, collaborative representations) and classical machine learning models (e.g., Isolation Forest) have significantly advanced HAD, they often come with limitations. Representation-based methods frequently require manual feature engineering, rely on strong assumptions about data structure, and require extensive parameter tuning. Similarly, while classical machine learning models offer scalability and unsupervised capabilities, they can rely on predefined distance metrics or random feature selection, which may limit their effectiveness in detecting subtle anomalies within complex hyperspectral datasets where intricate spatial and spectral relationships are important.

To address these challenges, recent advancements in deep learning-based approaches have gained considerable attention. Unlike classical statistical and tree-based methods, deep learning models are capable of automatically learning hierarchical features, effectively capturing spatial-spectral patterns from raw data and thus improving anomaly detection accuracy. This inherent ability eliminates the need for extensive manual feature engineering, leading to more adaptive, robust, and generalizable solutions for hyperspectral anomaly detection.

Based on current research trends in HAD, deep learning architectures can be broadly categorised into several approaches, including convolutional Neural Network(CNN) based models, reconstruction-based models, attention-driven networks, diffusion-based methods, generative models, and ensemble strategies. To provide a structured overview of specific implementations within these categories, 
Table \ref{tab:architectures} summarizes key deep learning models developed for HAD. For each listed model, it highlights its primary objective, the loss function used for optimisation, and the activation functions employed within its architecture, grouped by its core deep learning approach. This table serves as a quick reference to the diverse methodologies within deep learning for HAD.

\begin{table*}[htbp!]
\caption{Comparison of Key Deep Learning Models for Hyperspectral Anomaly Detection}
\label{tab:architectures}
\centering
\footnotesize
\setlength{\tabcolsep}{4pt} 
\begin{tabularx}
{\textwidth}{l >{\RaggedRight\arraybackslash}X >{\RaggedRight\arraybackslash}X >{\RaggedRight\arraybackslash}X >{\RaggedRight\arraybackslash}X @{}}

\toprule
\textbf{Approach Type} & \textbf{Model} & \textbf{Primary Objective} & \textbf{Loss Function} & \textbf{Activation Function} \\
\midrule

\textbf{Reconstruction}
& DSLF \cite{10309883} & Background Recon. & SRL \& SAD & N/A \\
\textbf{-based}& Auto-AD \cite{9382262} & Background Recon. & Adaptive Weighted Loss & Leaky ReLU \\
& RGAE \cite{9494034} & Background Recon. \& Regularization & $\ell_{2,1}$ Norm \& Graph Reg. & Sigmoid \\
& HSAE \cite{10571989} & Feature Extraction/Recon. & Recon \& Distribution Loss & ReLU \\
& TDD \cite{10506667} & Mask Generation \& Image Recon. & Weighted Cross-Entropy Loss & ReLU, Sigmoid \\
& LREN \cite{Jiang_Xie_Lei_Jiang_Li_2021} & Low Rank Est. \& Background Supp. & Recon. \& Density estimation & Tanh \\ 
\midrule

\textbf{GAN-based}
& HADGAN \cite{8972475} & Discriminative Anomaly Learning & Recon, \& Shrink constraint Loss & Leaky ReLU \\
& GAN-RX \cite{arisoy_gan-based_2021} & Synthetic HSI Gen. for AD & Adversarial (CE) \& $\ell_1$ Recon. Loss & Leaky ReLU, Sigmoid \\
\midrule

\textbf{CNN-based}
& CNND \cite{7875485} & Supervised Anomaly Detection & Categorical Cross-Entropy & Sigmoid \\
\midrule

\textbf{Attention-driven}
& GT-HAD \cite{lian_gt-had_2024} & Attention-based Feature Extraction & Mean Squared Error (MSE) & GELU \\
& AETNet \cite{10073635} & General Anomaly Enhancement & MSGMS Loss & Swin Transformer Block \\
\midrule

\textbf{Diffusion-based}
& BSDM \cite{ma_bsdm_2023} & Background Suppression \& Noise Est. & $\ell_2$ Norm on Estimated Noise & Tanh \\
& DWDDiff \cite{10858750} & Spectral Diffusion \& Background Est., Anomaly Synthesis & Recon. \& Abudance & ReLU \\
\midrule

\textbf{Ensemble Methods}
& GE-AD\cite{jimaging10060131} & Greedy Search on heterogeneous methods &N/A  &N/A  \\
\bottomrule
\end{tabularx}
\footnotesize
\raggedright

\textbf{Abbreviations}: AD - Anomaly Detection; HSI - Hyperspectral Image; Recon. - Reconstruction; Reg. - Regularization; Est. - Estimation; Supp. - Suppression; Gen. - Generation; CE - Cross-Entropy.
\textbf{Loss/Activation Not Specified}: Specific loss/activation functions were not explicitly detailed in the referenced papers denoted by N/A.

\end{table*}

\subsubsection{Convolutional Neural Networks}
 
Convolutional Neural Networks (CNNs) represent the class of feed-forward neural networks characterised by their hierarchical structure, which consists of convolutional layers, pooling layers, and fully connected layers. These networks exploit spatially local correlations within data by enforcing local connectivity patterns between neurons in adjacent layers. This makes CNN  well-suited for ROBUST feature extraction in hyperspectral anomaly detection \cite{https://doi.org/10.1155/2015/258619}, enabling them to learn relevant spatial features.

One notable approach, by the work in \cite{7875485}, employs deep CNNs for HAD by utilising a reference image scene with labelled samples. This method leverages the discriminative power of CNNs to measure the similarity or dissimilarity between image samples, thereby enhancing anomaly detection accuracy through a learned feature space.

However, relying solely on CNNs for anomaly detection can often yield suboptimal results. While excellent at local feature learning, conventional CNN architectures, especially fully connected layers, may fail to fully capture global spatial dependencies in hyperspectral images or to distinguish subtle anomalous patterns from complex backgrounds. To address these limitations, CNNs are frequently integrated with other deep learning architectures, such as autoencoders for feature refinement and robust background reconstruction, or with recurrent layers (like LSTMs) to capture sequential spectral dependencies, leading to more robust and comprehensive anomaly detection models.

\subsubsection{Reconstruction-based Models}
Reconstruction-based models capture nonlinear characteristics and operate on the principle of background reconstruction in hyperspectral images. Since anomalies are difficult to reconstruct, their presence can be identified by comparing the reconstructed image with the background. Autoencoders have been widely used for this purpose, as they can automatically capture abstract and hierarchical features while detecting pixels that deviate from the majority in the spectral domain. However, classical autoencoders are highly susceptible to noise, leading to modifications that enhance their robustness for HAD applications.

Several notable variations of autoencoders have been introduced to address these limitations. AUTO-AD \cite{9382262}, for example, employs a fully convolutional autoencoder with skip connections to reconstruct the background image while suppressing anomalies, which are typically small and sparse. Another variation, the robust autoencoder \cite{9494034}, incorporates spatial features and utilises $l_{2,1}$
normalisation to reduce noise sensitivity while implementing segmentation-based graph regularisation. However, this method can be computationally expensive due to its reliance on supergraph-based superpixel segmentation, which may result in some loss of spatial information. To further improve background regularisation, a hyperbolic space-based autoencoder \cite{10571989} was developed, using hyperbolic space representation to enhance reconstruction and anomaly detection performance. These advancements in reconstruction-based models have significantly improved hyperspectral anomaly detection by enhancing feature learning, noise robustness, and spatial information preservation, but still have difficulty in constructing generalised algorithms for different hyperspectral data.

\subsubsection{Diffusion Models}
Diffusion models are a type of generative model inspired by nonequilibrium thermodynamics. They generally consist of a forward diffusion process, where noise is incrementally added to real data, and a reverse diffusion process, where noise is gradually removed to recover the original data.

In the forward process, a clean data sample \( \mathbf{x}_0 \) is converted to a noisy version \( \mathbf{x}_t \) using:
\[
\mathbf{x}_t = \sqrt{\bar{\alpha}_t} \, \mathbf{x}_0 + \sqrt{1 - \bar{\alpha}_t} \, \boldsymbol{\epsilon}_t, \quad \boldsymbol{\epsilon}_t \sim \mathcal{N}(0, \mathbf{I})
\]
where \( \bar{\alpha}_t = \prod_{i=0}^{t} (1 - \beta_i) \), and \( \beta_i \in (0, 1) \) is a predefined noise schedule. As \( t \) increases, \( \mathbf{x}_t \) becomes increasingly noisy and approximates a standard Gaussian distribution. In the reverse process, a neural network \( \hat{\boldsymbol{\epsilon}}_\theta(\mathbf{x}_t, t) \), often U-Net-based, is trained to predict and remove the noise by minimizing:
\[
\mathcal{L} = \mathbb{E}_{t, \mathbf{x}_0, \boldsymbol{\epsilon}} \left\| \boldsymbol{\epsilon}_t - \hat{\boldsymbol{\epsilon}}_\theta(\sqrt{\bar{\alpha}_t} \mathbf{x}_0 + \sqrt{1 - \bar{\alpha}_t} \boldsymbol{\epsilon}, t) \right\|^2
\]

Diffusion models have recently gained attention due to their ability to model complex data distributions. In HAD, models often struggle to separate anomalies from complex backgrounds, leading to poor generalisation. To address this limitation, BSDM (Background Suppression Diffusion Model) \cite{ma_bsdm_2023} was developed — the first approach to introduce diffusion models for HAD. BSDM treats the background as noise and suppresses it using a learned pseudo-background noise distribution. Unlike standard Gaussian noise, BSDM constructs pseudo-background noise that mimics the actual background distribution. This innovation suggests that diffusion models could revolutionise HAD.

DWDDiff \cite{10858750} further extends this concept by introducing a spectral diffusion model and a dual-window diffusion process, yielding high spatial and spectral precision in background estimation and target detection, respectively.

\subsubsection{Attention-Driven Models}  
Attention-driven models use the power of transformers to enhance discrimination learning by capturing content similarity more effectively \cite{10042179}. Transformer-based architectures utilise self-attention mechanisms to model long-range dependencies, enabling better contextual understanding and anomaly detection in hyperspectral imagery.  

Recently, several hyperspectral anomaly detection models have been developed. GT-HAD\cite{lian_gt-had_2024} introduces a gated transformer-based approach that separates background and anomalies using a dual-branch network with an adaptive gating unit. AETNet\cite{10073635} utilises a vision transformer (ViT)-based model\cite{dosovitskiy2021imageworth16x16words} that requires only one-time training, significantly improving generalization across different datasets. TDD \cite{10506667} integrates a U-Net architecture\cite{ronneberger2015unetconvolutionalnetworksbiomedical} with both global and local self-attention mechanisms to enhance anomaly discrimination.  

While attention-based models demonstrate superior feature extraction capabilities, they also introduce challenges such as high computational costs due to the quadratic complexity of self-attention operations. To mitigate this, Mamba-based state-space models \cite{He_2024} \cite{Huang_2024} have been explored in hyperspectral classification. However, their potential in anomaly detection remains an open research area, requiring further investigation to improve efficiency and scalability.

\subsubsection{GAN-based Models}
The ability to model complex data distribution has encouraged the use of Generative Adversarial models in hyperspectral anomaly detection. GAN-based approaches consist of a generator and a discriminator, where the generator synthesises hyperspectral images, and the discriminator distinguishes between real and generated samples. One such approach GAN-RX \cite{arisoy_gan-based_2021} generates synthetic hyperspectral images and subtracts them from the original image to obtain a spectral difference image. The anomaly scores are then computed using the RX-AD algorithm, which enhances anomaly-background separability. The weights of the generator and  the discriminator networks are trained according to the following objective functions
\begin{equation*}
    \begin{array}{c} 
    {\mathcal{L}_{GAN}}(G,D) = {{\text{E}}_{\text{s}}}[\log (D({\text{s}}))] + \\ {{\text{E}}_{\text{s}}}[\log (1 - D(G({\text{s}})))], \end{array} 
\end{equation*} where where $\mathbf{s} \in \mathbb{R}^L$ is the real spectral vector.
Similarly, Sparse-GAN \cite{li_sparse_2021} learns a discriminative latent reconstruction, ensuring small reconstruction errors for background pixels while producing large errors for anomalous pixels, effectively highlighting anomalies.  

While GAN-based models offer powerful feature learning capabilities, challenges remain in stabilising training and improving generalisation across diverse hyperspectral datasets. 

\subsubsection{Ensemble-based models}  
Different hyperspectral anomaly detection algorithms model the background in various ways, making it challenging to capture all background constraints across diverse scenarios. To address this, ensemble-based approaches have been developed, combining multiple models to improve anomaly detection performance in hyperspectral imagery.  

One such method is GE-AD\cite{jimaging10060131}, a greedy ensemble-based anomaly detector that utilises a greedy algorithm to combine multiple detection techniques, including AED,LSAD\cite{DU2016115}, CSD, RX, KIFD, LSUN,LSAD-CR-IDW\cite{article} and RSORAD. By using the strengths of different models, GT-HAD enhances anomaly detection accuracy and robustness across complex environments.  

While ensemble models significantly improve detection performance, they introduce challenges such as increased computational complexity, model selection, and parameter tuning. Future research should focus on optimising ensemble strategies to achieve a balance between computational efficiency and detection accuracy, making them more scalable for real-world applications.

Each category of methods has distinct advantages depending on the application context, data availability, and computational constraints. A detailed comparison and analysis of these approaches is presented in later sections.
\begin{table*}[htbp!]
\caption{Advantages and Limitations of Different HAD Algorithms}
\label{tab:ago-comparison}
\centering
\footnotesize
\begin{tabularx}{\textwidth}{@{} l l X X @{}}
\toprule
\textbf{Category} & \textbf{Algorithm} & \textbf{Advantages} & \textbf{Limitations} \\
\midrule

\multirow{5}{*}{\textbf{Statistical} } 
& RX & Fast computation time. Performs well in simple scenes. 
& Performs poorly in complex scenes when the Gaussian distribution assumption is violated. \\

\textbf{Methods}& KRX & Transforms a nonlinear method into a linear one in a high-dimensional feature space. 
&Computationally intensive; neglects spatial context by primarily focusing on spectral data. \\

& LRX & Uses a local dual window to capture local background statistics and eliminate anomalous pixels. 
& High computational cost; sensitive to window size. \\

& Subspace RX & Uses the principal components of the covariance matrix to suppress background features. 
& Poor performance in noisy environments \\

& FrFT-RX & Handles non-stationary noise effectively. 
& Suboptimal performance in certain noisy conditions; parameter sensitivity \\
\midrule

\multirow{4}{*}{\textbf{Representation}} 
& CRD & Does not assume a random background distribution; represents each background pixel using surrounding pixels. 
& Sensitive to noise and window size; assumes uniform band contribution. \\

\textbf{-Based}& LRASR & utilises both local and global HSI information. 
& High computational requirements (matrix decomposition, SVD); sensitive to noise. \\

& PTA & Employs normalization for robust low-rank approximation; noise-robust. 
& High computational cost. \\

& RPCA & Works effectively for detecting isolated pixels. 
& High false alarm rate. \\
\midrule

\multirow{2}{*}{\textbf{Classical ML}} 
& KIFD & Accurate outlier identification; scalable to large datasets.
& Struggles with local anomaly detection; sensitive to parameter tuning (e.g., number of trees). \\

& SVDD & Efficient at finding minimum enclosing boundaries in feature space. 
& Sensitive to kernel choice and parameter settings; limited in capturing complex structures in high-dimensional data. \\
\midrule

\multirow{3}{*}{\textbf{Deep Learning }} 
& Auto-AD & Uses adaptive weighted loss to suppress background and approximate local features. 
& Complex reconstruction network; risk of overfitting. \\

\textbf{Methods}& CNND & Utilises differences between adjacent pixel pairs to effectively capture local variations 
& Requires ground truth labels for training; high computational complexity. \\

& GT-HAD & Utilises attention mechanism, content similarity and residual diffusion for roboust background reconstruction.
& Struggles with detecting anomalies near boundaries; computationally intensive. \\
\bottomrule
\end{tabularx}
\end{table*}

\section{Current Trends in Hyperspectral Data Preparation and Noise Handling}~\label{sec:trends}  
\subsection{Input Data Types}  
The processing of hyperspectral images for anomaly detection can vary based on how the input data is structured. The choice of input data type is crucial, as it dictates the balance between preserving spatial and spectral information, managing computational costs, and aligning with the underlying model's design.
\par Whole image-based processing takes the entire hyperspectral image as input. This approach preserves global spatial and spectral relationships, making it suitable for deep learning models that analyse overall image-level patterns. For example, autoencoders and robust autoencoders use the entire image as input to learn holistic features for anomaly detection.

\par Patch-based processing divides the image into smaller patches instead of using the whole image. This method is useful for detecting localised anomalies and reducing computational complexity. he TDD model extracts patches from images to simulate anomalies and trains its architecture using these smaller regions. Similarly, models like GT-HAD utilise small pixel-based cubes spanning multiple spectral bands (effectively 3D patches centred on pixels) to capture both local spatial and spectral features efficiently.

\par Pixel-based processing treats each pixel independently, considering only its spectral vector across all bands. This method is computationally efficient and has historically been popular in statistical (RX Algorithm), which processes individual pixels separately and classical machine learning approaches. It is useful when spatial information is less important than precise spectral characterisation for anomaly detection.

Superpixel-based processing involves grouping similar and spatially contiguous pixels into larger, irregular regions called superpixels. This method aims to reduce noise and redundancy by aggregating local pixel information while largely preserving spatial boundaries, making it useful for segmentation-based deep learning models.  For example, the HSAE converts images into superpixels and maps extracted features into hyperbolic space for regularisation. Similarly, some modified isolation forest-based models, such as SSIIFD, also utilise superpixel-based inputs.

\subsection{HSI Pre-processing}
\subsubsection{Noise Handling Techniques in Hyperspectral Imaging}  

In real-world hyperspectral imaging, noise arises due to environmental variations, sensor limitations, and data acquisition processes that can significantly degrade the performance of anomaly detection algorithms. Consequently, denoising techniques have become an essential preprocessing step to enhance the quality of extracted features. However, it is important to understand that crude denoising can unintentionally erase important image details, such as geometrical structures and anomaly edges, leading to a loss of detection accuracy.  

To address this, additive noise such as Gaussian noise, salt-and-pepper noise, deadlines, and stripes is  sometimes intentionally introduced during training. This data augmentation strategy enables algorithms to learn noise patterns, thereby preventing overfitting to specific artifacts and improving model robustness. A more advanced paradigm moves beyond separate denoising and anomaly detection steps, with modern frameworks integrating joint denoising and anomaly detection \cite{10621578}. In these systems, the outputs of one process iteratively refine the other, leading to an improvement in performance. 

Since anomalies in hyperspectral images are rare and lack prior information, various feature extraction and denoising techniques are employed to enhance detection accuracy. These can broadly be categorised into model-driven and data-driven approaches:
\paragraph{Model-Driven Denoising Methods}
These techniques rely on explicit mathematical models of noise or image properties. Guided filtering, for instance, helps smooth images while preserving spatial consistency \cite{9426593}, making anomalies more distinguishable. Low-Rank Representation is also effective in suppressing noise by modelling the background as a structured low-rank subspace while isolating anomalies as sparse outliers \cite{9426593}. Other methods use techniques like Principal Component Analysis (PCA) and Gabor filters for noise removal and discriminative feature retention, with strategies such as dictionary pruning \cite{7119558} or isolation-based models \cite{10553340}. A common limitation of these model-driven methods is that they often rely on manually optimised parameters and are designed to eliminate noise of a fixed intensity or specific type, which can result in difficulties when tackling the complex and variable noise distributions encountered in real-world HSI.

\paragraph{Data-Driven (Deep Learning) Denoising Methods}
Modern deep learning-based methods are also data-driven methods for HSI denoising, learning complex noise patterns directly from data.  Methods such as HyADD \cite{10621578} use spatial–spectral total variation (SSTV) regularisation along with an antinoise dictionary, ensuring that anomaly detection and denoising work together iteratively to improve robustness. HSI-DeNet \cite{8435923} utilises deep convolutional neural networks with a series of multichannel 2D filters in HSI denoising. Similarly, HSI-SDeCNN \cite{8913713} presents another denoising method using a single CNN architecture. QRNN3D \cite{9046853} focuses on deep denoising by introducing QRNN into the 3D U-net structure capable of handling sequential spectral information. Cai, Y. et.al \cite{10.1007/978-3-031-19790-1_41} utilised a sparse Transformer to embed sparsity of HSIs into deep learning for HSI reconstruction and denoising. 
\par Despite these recent advancements, noise removal in hyperspectral images remains an ongoing and critical research area, particularly due to the complex and often unknown distribution of noise, especially within the spectral domain. Further studies are needed to develop more efficient, adaptive, and scene-specific denoising techniques that can robustly enhance anomaly detection while preserving essential image features.

\subsubsection{Feature Extraction in Hyperspectral Imaging}  
Feature extraction is an important step in HAD, serving to transform raw high-dimensional spectral data into more compact, informative, and discriminative representations suitable for classification or anomaly identification. Various methodologies have been developed for effective feature extraction, which can be broadly categorised based on their underlying principles: statistical, representation, and deep learning techniques.

\paragraph{Statistical Methods}  

Statistical methods are widely employed for dimensionality reduction and feature extraction, focusing on variance preservation and decorrelation. PCA \cite{7994698} is a technique that extracts dominant spectral features by transforming data into a lower-dimensional subspace while preserving maximal variance. Kernel PCA \cite{8989981} extends this by implicitly mapping data into a high-dimensional feature space via a kernel function, allowing for the extraction of more complex, non-linear feature transformations. These algorithms excel at capturing spectral components. Beyond dimensionality reduction, other statistical approaches include spectral band selection, where redundant, noisy, or less informative bands (e.g., water absorption bands) are identified and removed to enhance detection accuracy and reduce computational load.

\paragraph{Representation-Based Methods}  

Building upon the principles of signal reconstruction, sparse and low-rank representation models are commonly utilised for feature extraction in hyperspectral background modelling. Low-Rank Representation models the background as a structured, low-dimensional subspace, from which anomalies naturally emerge as sparse deviations or outliers. Similarly, dictionary learning techniques \cite{SU2020195},like collaborative and robust sparse representation, are applied to learn overcomplete dictionaries. These dictionaries facilitate the isolation of meaningful spectral signatures, thereby improving feature discriminability while effectively suppressing noise.

\paragraph{Deep Learning-Based Methods}  

Recent advancements in deep learning have revolutionised feature extraction in HAD by enabling automated learning of hierarchical and abstract representations. Autoencoders and their various architectures, such as stacked autoencoders and graph-based autoencoders, are widely used to learn compressed, robust feature representations from hyperspectral data, often by minimising reconstruction error. Attention mechanisms (e.g., in Transformers) and CNNs are extensively employed to capture intricate spatial-spectral dependencies, utilising their ability to learn local and global patterns. Furthermore, specialised deep models, such as tensor-based feature extraction networks, can incorporate operations like Fourier transforms \cite{Zhang19052021} to enhance the representation of spectral features. Generative Adversarial Networks (GANs) \cite{WANG2023109795} have also been explored for feature extraction, primarily by learning to synthesise realistic spectral distributions, thereby assisting in robust background modelling and subsequent anomaly identification.

\paragraph{Integrated and Hybrid Approaches}
Recognising the complementary strengths of different paradigms, many modern frameworks adopt integrated or hybrid techniques to achieve superior feature extraction. This often involves combining deep learning models with classical statistical methods; for instance, background estimation networks might use deep learning to process data that has undergone initial dimensionality reduction via PCA. Other approaches in classical machine learning (and sometimes integrated with deep learning) utilise Gabor filters for extracting spatial features, which can enhance anomaly contrast and preserve structural integrity. Additionally, entropy-based feature selection methods are sometimes employed to quantitatively identify the most informative spectral bands, further optimising the input for various anomaly detection algorithms.

Overall, feature extraction in hyperspectral imaging continues to evolve, with emerging techniques focusing on enhancing spatial-spectral consistency, improving robustness to noise and irrelevant features, and integrating sophisticated domain-specific knowledge to achieve more accurate and reliable anomaly detection performance. 

\section{Comparison Analysis}~\label{sec:analysis}
\subsection{Datasets for Hyperspectral Anomaly Detection}
\begin{table*}[htbp!]
\centering
\footnotesize 
\setlength{\tabcolsep}{2 pt}
\renewcommand{\arraystretch}{1.3}
\caption{Comparison of Popular Hyperspectral Anomaly Detection Datasets}
\label{tab:datasets}

\begin{tabular}{c l l c c c c  l l}
\toprule
\textbf{ID} & \textbf{Dataset} & \textbf{Sensor} & \textbf{Spectral} & \textbf{Spatial} & \textbf{Bands} & \textbf{Res.} & \textbf{Background} & \textbf{Anomalies} \\
 &  &  & \textbf{Range (nm)} & \textbf{Size} &  & \textbf{(m)} &  &  \\
\midrule
1 & \href{https://doi.org/10.1109/TGRS.2015.2493201}{San Diego} & AVIRIS & 370--2510 & 100$\times$100 & 186 & 3.5 & Airport & Airplanes \\
2 & \href{http://www.tec.army.mil/Hypercube}{HYDICE} & HYDICE & 400--2500 & 800$\times$100 & 175 & 1.56 & Urban area & Roofs, cars \\
3 & \href{https://doi.org/10.1007/s10618-010-0182-x}{Cri} & Nuance Cri & 650--1100 & 400$\times$400 & 46 & -- & Grass & Ten rocks \\

4 & \href{https://www.ehu.eus/ccwintco/index.php/Hyperspectral_Remote_Sensing_Scenes}{Salinas}  & AVIRIS & 400--2500 & 512$\times$217 & 224 & 3.7 & Salinas Valley & Rooftops \\
5 & \href{https://doi.org/10.1109/TGRS.2022.3144192}{Pavia University} & ROSIS & 430--860 & 224$\times$423 & 102 & 1.3 & Pavia, Italy & Vehicles \\

6 & \href{https://github.com/sxt1996/AVIRIS-1}{AVIRIS-1} & AVIRIS & 370--2510 & 100$\times$100 & 189 & 3.5 & San Diego & Airplanes \\

7 & \href{https://github.com/sxt1996/AVIRIS-2}{AVIRIS-2} & AVIRIS & 370--2510 & 128$\times$128 & 189 & 3.5 & San Diego & Airplanes \\

\midrule
\multicolumn{9}{c}{\textbf{\href{http://xudongkang.weebly.com/data-sets.html}{ABU-Airport Subcategories}}} \\
\midrule
8.1 & Los Angeles-1 & AVIRIS & 430--860 & 100$\times$100 & 205 & 7.1 & Airport & Airplanes \\
8.2 & Gulfport & AVIRIS & 400--2500 & 100$\times$100 & 191 & 3.4 & Gulfport & Airplanes \\
8.3 & Los Angeles-2 & AVIRIS & 430--860 & 100$\times$100 & 205 & 7.1 & Airport LA & Airplanes \\
\midrule

\multicolumn{9}{c}{\textbf{ABU-Urban Subcategories}} \\
\midrule
9.1 & Texas Coast-1 & AVIRIS & 400--2500 & 100$\times$100 & 204 & 17.2 & Urban Texas & Buildings \\

9.2 & Texas Coast-2 & AVIRIS & 400--2500 & 100$\times$100 & 207 & 17.2 & Urban Texas & Buildings \\
9.3 & Gainesville & AVIRIS & 400--2500 & 100$\times$100 & 191 & 3.5 & Gainesville & Buildings \\
9.4 & Los Angeles-3 & AVIRIS & 430--860 & 100$\times$100 & 205 & 7.1 & Urban LA & Houses \\
9.5 & Los Angeles-4 & AVIRIS & 430--860 & 100$\times$100 & 205 & 7.1 & Urban LA & Houses \\

\midrule
\multicolumn{9}{c}{\textbf{ABU-Beach Subcategories}} \\
\midrule
10.1 & Cat Island & AVIRIS & 400--2500 & 150$\times$150 & 188 & 17.2 & Beach & Plane \\
10.2 & Bay Champagne & AVIRIS & 400--2500 & 100$\times$100 & 188 & 4.4 & Coastal region & Watercraft \\
10.3 & San Diego & AVIRIS & 400--2500 & 100$\times$100 & 193 & 7.5 & Beach near SD & Boats, buildings \\
10.4 & Pavia & ROSIS-03 & 430--860 & 150$\times$150 & 102 & 1.3 & Pavia coastal & Vehicles on bridge \\
\midrule
11 & \href{https://zhaoxuli123.github.io/HAD100/}{HAD100} & AVIRIS & variable & 64$\times$64 & 276/162 & -- & variable & Vehicles, buildings \\
12&\href{http://rsidea.whu.edu.cn/resource_WHUHiriver_sharing.htm}{WHU-Hi-River} & Headwall Nano & 400--1000 & 105$\times$168 & 135 & 6 & WHU-Hi-River & Plastic plates, panels \\
13 & \href{https://doi.org/10.3390/rs11111318}{HySpex} & AVIRIS & 415--2508 & 170$\times$130 & 288 & 0.73 & Xuzhou, China & Colored steel roofs \\
14 & \href{https://doi.org/10.1109/TGRS.2022.3144192}{El Segundo} & AVIRIS & 366--2496 & 250$\times$300 & 224 & 7.1 & Oil refinery & Roofs, vehicles \\

\bottomrule
\end{tabular}
\end{table*}

In this section, we present an overview of publicly available hyperspectral datasets commonly utilised for benchmarking anomaly detection algorithms. Table~\ref{tab:datasets} summarizes key characteristics of each dataset, including the sensor type, spectral range, spatial dimensions, number of spectral bands, spatial resolution, and descriptions of background and anomaly types.

\par A majority of these datasets are acquired using the Airborne Visible/Infrared Imaging Spectrometer (AVIRIS) sensor, including the San Diego and ABU datasets. Other prominent sensors featured in these collections include the HYDICE sensor for the HYDICE dataset, the Nuance Cri sensor for the Cri dataset, and the ROSIS sensor for the Pavia University dataset. The ABU datasets, a widely used collection, are further categorised into three distinct scene types: \textit{Abu-Urban}, \textit{Abu-Beach}, and \textit{Abu-Airport}. For each dataset, the spatial size denotes the image dimensions in pixels (height × width), while the spatial resolution (in meters per pixel) quantifies the ground sampling distance. The number of bands indicates the total spectral channels captured, and the spectral range specifies the wavelength coverage in nanometers. These HAD benchmarking datasets exhibit relatively low spatial resolution but offer a rich spectral range across the electromagnetic spectrum. Beyond these established real-world collections, recent trends include the development of synthetic datasets, where anomalies are artificially introduced into real scenes by masking while preserving spectral properties \cite{10506667}. Additionally, large-scale real-scene collections, such as HAD100 \cite{10073635}, have been introduced to foster more robust and generalised hyperspectral anomaly detection research.

\subsection{Evaluation Metrics}
\subsubsection{ROC Curves and AUC (Receiver Operating Characteristic and Area Under the Curve)}
The ROC curve is a widely recognised tool for visualising and evaluating the performance of anomaly detection algorithms and binary classifiers \cite{FAWCETT2006861}. An ROC graph illustrates the inherent trade-off between the true positive rate (detection probability) and the false positive rate (false alarm rate) across various classification thresholds.

The AUC is a scalar metric that quantifies the overall performance of a classifier, summarised from its ROC curve. It represents the portion of the unit square occupied by the ROC curve, with values ranging from 0 to 1.  AUC is equivalent to the probability that a randomly chosen positive instance (anomaly) is ranked higher by the classifier than a randomly chosen negative instance (background).

For HAD, performance is assessed using key rates: the probability of detection ($P_D$), which is the true positive rate, and the probability of false alarms ($P_F$), representing the false positive rate. These are calculated as follows:
\begin{equation}
    P_D = \frac{TP}{TP + FN}
    \label{eq:pd}
\end{equation}
\begin{equation}
    P_F = \frac{FP}{TN + FP}
    \label{eq:pf}
\end{equation}
where $TP$ (True Positives) refers to target pixels correctly classified as anomalies, $FN$ (False Negatives) represents missed target pixels, $FP$ (False Positives) indicates background pixels incorrectly classified as anomalies, and $TN$ (True Negatives) are correctly classified background pixels.

The standard 2D AUC value, assessing the overall trade-off between $P_D$ and $P_F$, is computed using the trapezoidal rule:
\begin{equation}
    AUC = \frac{1}{2} \sum_{l=1}^{p-1} (P_F^{l+1} - P_F^l)(P_D^{l+1} + P_D^l)
    \label{eq:auc_2d}
\end{equation}
where $(P_F^l, P_D^l)$ represents the $l^{th}$ point on the discrete ROC curve, and $p$ is the total number of points. A higher AUC value generally indicates superior classifier performance.

In the context of HAD,  when considering the influence of the detection threshold ($\tau$), 3D ROC curves can be employed. From these, specialized AUC metrics considering the threshold dimension are derived: $AUC_{D,\tau}$ measures the classifier's ability to achieve high detection probability at different threshold values, and $AUC_{F,\tau}$ evaluates the classifier's performance concerning the false positive rate as the threshold varies. These are calculated as:
\begin{equation}
    AUC_{D,\tau} = \frac{1}{2} \sum_{l=1}^{p-1} (\tau^{l+1} - \tau^l) (P_D^{l+1} + P_D^l)
    \label{eq:aucd_tau}
\end{equation}
\begin{equation}
    AUC_{F,\tau} = \frac{1}{2} \sum_{l=1}^{p-1} (\tau^{l+1} - \tau^l) (P_F^{l+1} + P_F^l)
    \label{eq:aucf_tau}
\end{equation}
Using these threshold-dependent AUC values, the Signal-to-Noise Probability Ratio (SNPR), a metric valuable in HAD, is calculated as:
\begin{equation}
    SNPR = \frac{AUC_{D,\tau}}{AUC_{F,\tau}}
    \label{eq:snpr}
\end{equation}
For performance interpretation, a higher value of AUC ($P_D, P_F$) or $AUC_{D,\tau}$ generally indicates better anomaly detection capabilities. Conversely, a lower value of $AUC_{F,\tau}$ suggests more effective background suppression. A larger SNPR value is also indicative of superior overall anomaly detection performance, particularly reflecting the balance between anomaly detection and background clutter rejection.

\subsubsection{Box-Whisker Plot}
The Box-Whisker Plot, is a Statistical Separability Map, serving as a powerful visualization tool to assess the discriminability between background and anomalous pixels in hyperspectral anomaly detection \cite{10042179}. As illustrated in Figure~\ref{fig:box}, the distribution of anomaly scores for anomalous pixels is represented by one box (red box), while the distribution for background pixels is shown by another (a blue box). The line within each box indicates the median value of its respective score distribution. 

The effectiveness of an anomaly detection algorithm is quantifiable through two key aspects of this plot. Firstly, the interval between the anomaly and background boxes quantifies their separability: a larger, non-overlapping gap signifies superior distinction between anomalous and background pixels. Secondly, the height of the background box reflects the degree of background suppression: a shorter box for background pixels suggests a tighter distribution of background scores around its median, indicating stronger suppression and, consequently, improved anomaly-background separation \cite{hou2022spectralspatialfusionanomalydetection}. Thus, a clear separation between the boxes indicates high detectability of anomalies, while overlapping regions point to challenges in distinguishing targets from the background.

\subsubsection{Square Error Ratio (SER) and Area Error Ratio (AER)}
The Square Error Ratio (SER) \cite{10309883} is the quantitative metric that measures how far off the predictions are from the ground truth. Anomaly pixels are considered 1 and 0 for background pixel. It calculates the sum of squared differences between the ground truth and the predicted values for both background and anomaly pixels. It is mathematically expressed as:

\begin{equation*}
\text{SER} = \frac{\sum_{i=1}^{N_a} \left( \text{Pre}_i - 1 \right)^2 + \sum_{j=1}^{N_b} \left( \text{Pre}_j - 0 \right)^2}{N} \times 100 \quad \tag{16}
\end{equation*}

where \( N_a \) and \( N_b \) represent the number of anomaly and background pixels, respectively, and \( N = N_a + N_b \) denotes the total number of pixels in the hyperspectral image (HSI). Here, \( \text{Pre}_i \) and \( \text{Pre}_j \) refer to the predicted outlier scores for anomaly and background pixels. A lower SER indicates a closer match between predicted scores and the ground truth, which reflects higher detection precision. This metric is used when ground-truth pixel-level labels are available.

The Area Error Ratio (AER) evaluates the quality of anomaly detection by measuring the tradeoff between false positives and detection accuracy over a range of thresholds. classical ROC curves plot the true positive rate (TPR) against the false positive rate (FPR) for various thresholds, and their AUC (area under the curve) is commonly used as a scalar performance measure. However, ROC curves lack the ability to assess how confidently anomalies are separated from background pixels, as they are insensitive to the actual predicted score magnitudes.

To address this, a modified ROC representation is proposed, plotting both TPR and FPR against normalized threshold values derived from the predicted scores.The TPR curve should reach point (1, 1), indicating full detection at all thresholds.The FPR curve should remain close to zero across all thresholds.\cite{8082110}

In such conditions, the maximum achievable area under the TPR and FPR curves would be 1 and 0, respectively. The AER is defined as: 

\begin{equation}
    AER = \frac{\sum_{l=1}^{p} P_f^l}{\sum_{l=1}^{p} P_d^l}
\end{equation}
where \( p \) is the number of thresholds evaluated,\( P_f^l \) is the false positive rate (FPR) at threshold \( l \), \( P_d^l \) is the true positive rate (TPR) at threshold \( l \).

A lower AER implies that the algorithm maintains a low false alarm rate while achieving high detection across various thresholds.

\subsubsection{Mean Square Error (MSE)}
The Mean Square Error (MSE) quantifies the average squared difference between the estimated and actual values of a detection algorithm. This is mainly used in reconstruction-based algorithm where $X_i$ is denoted as original image and $\hat{X}_i$ as the reconstructed image.  It is computed as:

\begin{equation}
    MSE = \frac{1}{N} \sum_{i=1}^{N} (X_i - \hat{X}_i)^2
\end{equation}

where $X_i$ represents the actual value,$\hat{X}_i$ represents the estimated value and $N$ is the total number of samples.

A lower MSE indicates higher accuracy, as it means the predicted values are closer to the true values.

\subsection{Performance Comparison}  
In this survey, we comprehensively evaluate ten representative HAD algorithms across seventeen widely used benchmarking hyperspectral datasets. The evaluation primarily focuses on two key performance metrics: Area Under the ROC Curve (AUC), for detection accuracy, and execution time (in seconds), for computational efficiency. The diverse range of datasets, captured by various sensors with distinct spectral bands and scene characteristics, provides a robust environment for assessing model generalization. The algorithms selected span four main categories: statistical methods (RX, LRX), representation-based models (PTA, CRD), classical machine learning (KIFD), and deep learning-based techniques (Auto-AD, RGAE, TDD, LREN, GT-HAD).

The detailed quantitative results are summarised in Table~\ref{tab:metrics}, offering insights into the inherent trade-offs between detection accuracy and computational overhead across these diverse methodologies.

\begin{table*}[htbp!]
\centering
\caption{Performance Comparison of Hyperspectral Anomaly Detection Algorithms. Best Area Under Curve (AUC) and shortest execution time for each dataset and overall mean are highlighted in \textbf{bold}. Datasets are identified by their IDs, as further detailed in Table~\ref{tab:datasets}.}
\label{tab:metrics}
\footnotesize
\begin{tabular}{@{\hspace{3pt}}l@{\hspace{9pt}}c@{\hspace{9pt}}c@{\hspace{9pt}}c@{\hspace{9pt}}c@{\hspace{9pt}}c@{\hspace{9pt}}c@{\hspace{9pt}}c@{\hspace{9pt}}c@{\hspace{9pt}}c@{\hspace{9pt}}c@{\hspace{9pt}}c@{\hspace{9pt}}c@{\hspace{0pt}}}

\toprule
\textbf{ID} & \textbf{Metric} & \textbf{RX} & \textbf{LRX} & \textbf{CRD} & \textbf{PTA} & \textbf{KIFD} & \textbf{Auto-AD} & \textbf{RGAE} & \textbf{TDD} & \textbf{LREN} & \textbf{GT-HAD} \\
\midrule
\multirow{2}{*}{1} 
& AUC & 0.9403 & 0.9152 & 0.9637 & 0.9314 & 0.9913 & 0.9471 & \textbf{0.9920} & 0.9447 & 0.9704 & 0.9815 \\
& Time (s) & \textbf{0.28} & 15.89 & 15.74 & 21.68 & 36.86 & 8.66 & 78.82 & 1.52 & 38.41 & 10.20 \\
\midrule
\multirow{2}{*}{2} 
& AUC & 0.9857 & 0.9922 & 0.9942 & 0.8945 & \textbf{0.9965} & 0.9872 & 0.7602 & 0.7700 & 0.7001 & 0.8718 \\
& Time (s) & \textbf{0.22} & 9.66 & 11.61 & 15.86 & 24.02 & 9.53 & 66.00 & 0.76 & 50.06 & 7.08 \\
\midrule

\multirow{2}{*}{3} 
& AUC & 0.9989 & 0.7169 & 0.9899 & 0.4470 & 0.9942 & 0.5123 &\textbf{0.9983} & 0.9906 & 0.9976 & 0.9952 \\
& Time (s) & \textbf{0.18} & 5.19 & 198.52 & 97.00 & 242.27 & 29.23 & 608.82 & 11.70 & 557.93 & 290.13 \\
\midrule
\multirow{2}{*}{4} 
& AUC & 0.8073 & 0.9292 & 0.9674 & \textbf{0.9948} & 0.9930 & 0.9908 & 0.9291 & 0.2993 & 0.7665 & 0.9474 \\
& Time (s) & \textbf{0.41} & 26.72 & 26.10 & 37.96 & 52.27 & 7.54 & 166.91 & 1.57 & 137.59 & 19.57 \\
\midrule
\multirow{2}{*}{5} 
& AUC & 0.9538 & 0.9198 & 0.9389 & 0.9760 & 0.8054 & 0.9695 & 0.9042 & 0.2344 & 0.8977 & \textbf{0.9994} \\
& Time (s) & \textbf{0.57} & 6.19 & 30.22 & 25.40 & 68.54 & 22.68 & 210.82 & 0.63 & 66.64 & 30.01 \\
\midrule

\multirow{2}{*}{6} 
& AUC & 0.8866 & 0.7785 & \textbf{0.9911} & 0.9825 & 0.9879 & 0.7954 & 0.9886 & 0.9520 & 0.9757 & 0.9862 \\
& Time (s) & \textbf{0.30} & 14.96 & 16.31 & 22.83 & 48.59 & 8.71 & 80.98 & 2.77 & 60.55 & 10.54 \\
\midrule

\multirow{2}{*}{7} 
& AUC & 0.9181 & 0.8432 & 0.9115 & 0.8503 & 0.9327 & 0.9034 & 0.9392 & 0.9091 & \textbf{0.9342} & 0.8970 \\
& Time (s) & \textbf{0.38} & 29.33 & 26.96 & 36.93 & 44.90 & 9.27 & 203.60 & 1.39 & 101.71 & 21.26 \\
\midrule

\multirow{2}{*}{8.1} 
& AUC & 0.8221 & 0.9153 & 0.9227 & 0.8809 & 0.9383 & 0.9258 & 0.6389 & 0.8814 & 0.7068 & \textbf{0.9464} \\
& Time (s) & \textbf{0.36} & 19.35 & 16.72 & 21.79 & 46.57 & 7.54 & 263.99 & 1.08 & 129.14 & 14.17 \\
\midrule
\multirow{2}{*}{8.2} 
& AUC & 0.9526 & 0.7917 & 0.9675 & \textbf{0.9946} & 0.9824 & 0.9798 & 0.7461 & 0.5304 & 0.7774 & 0.9869 \\
& Time (s) & \textbf{0.28} & 16.06 & 15.63 & 24.00 & 43.82 & 2.01 & 77.87 & 1.33 & 49.51 & 9.92 \\
\midrule

\multirow{2}{*}{8.3} 
& AUC & 0.8404 & 0.8986 & 0.9758 & 0.9104 & 0.9728 & 0.8931 & 0.7483 & 0.9631 & 0.9078 & \textbf{0.9884} \\
& Time (s) & 1.77 & 16.35 & 14.69 & 21.14 & 43.51 & 3.93 & 84.81 & \textbf{1.29} & 106.16 & 17.72 \\

\midrule
\multirow{2}{*}{9.1} 
& AUC & 0.9907 & 0.9712 & 0.9694 & 0.9509 & 0.9083 & 0.9897 & 0.9821 & 0.1040 & 0.9834 & \textbf{0.9971} \\
& Time (s) & \textbf{0.29} & 21.16 & 16.62 & 24.92 & 42.71 & 3.38 & 84.57 & 1.24 & 163.69 & 10.87 \\

\midrule

\multirow{2}{*}{9.2} 
& AUC & 0.9946 & 0.9253 & 0.9428 & 0.8643 & 0.8608 & 0.9852 & \textbf{0.9994} & 0.1825 & 0.9823 & 0.9944 \\
& Time (s) & \textbf{0.32} & 19.96 & 12.19 & 21.50 & 41.77 & 3.33 & 258.27 & 1.11 & 96.08 & 14.21 \\
\midrule

\multirow{2}{*}{9.3} 
& AUC & 0.9513 & 0.9442 & 0.9235 & 0.9381 & \textbf{0.9923} & 0.9734 & 0.8205 & 0.9275 & 0.7992 & 0.9703 \\
& Time (s) & \textbf{0.24} & 15.08 & 12.10 & 19.16 & 45.03 & 7.36 & 238.39 & 3.55 & 55.12 & 14.42 \\
\midrule

\multirow{2}{*}{9.4} 
& AUC & 0.9887 & 0.8842 & 0.9619 & 0.7107 & 0.9787 & 0.9901 & \textbf{0.9948} & 0.8579 & 0.9252 & 0.9887 \\
& Time (s) & \textbf{0.23} & 17.86 & 12.67 & 20.77 & 39.68 & 3.74 & 254.03 & 3.63 & 101.34 & 14.30 \\
\midrule

\multirow{2}{*}{9.5} 
& AUC & 0.9692 & 0.9319 & 0.9413 & 0.9396 & \textbf{0.9867} & 0.8459 & 0.9569 & 0.8448 & 0.4369 & 0.9443 \\
& Time (s) & \textbf{0.25} & 17.95 & 12.60 & 21.18 & 40.92 & 9.54 & 117.50 & 2.76 & 95.73 & 14.49 \\
\midrule
\multirow{2}{*}{10.1} 
& AUC & 0.9807 & 0.9934 & \textbf{0.9980} & 0.9788 & 0.9898 & 0.9825 & 0.9393 & 0.2016 & 0.9310 & 0.9960 \\
& Time (s) & \textbf{0.64} & 34.18 & 35.65 & 53.15 & 57.11 & 8.63 & 186.60 & 1.40 & 70.14 & 33.84 \\
\midrule
\multirow{2}{*}{10.2} 
& AUC & 0.9999 & 0.9965 & \textbf{1.0000} & 0.9963 & 0.9978 & 0.9990 & 0.8662 & 0.2809 & 0.2387 & 0.9997 \\
& Time (s) & \textbf{0.23} & 14.74 & 12.90 & 20.40 & 44.55 & 6.43 & 240.93 & 1.12 & 89.60 & 13.91 \\
\midrule

\multirow{2}{*}{\textbf{Avg.}} 
& \textbf{AUC} & 0.9390 & 0.9013 & \textbf{0.9567} & 0.9141 & 0.9529 & 0.9273 & 0.8846 & 0.6468 & 0.8297 & \textbf{0.9733} \\
& \textbf{Time (s)} & \textbf{0.40} & 17.21 & 31.96 & 30.40 & 57.51 & 8.05 & 176.80 & \textbf{2.24} & 142.17 & 30.51 \\
\bottomrule
\end{tabular}
\end{table*}

\subsubsection{Result Discussion}
\begin{figure*}[htbp!]
    \centering
    \includegraphics[width=0.9\textwidth]{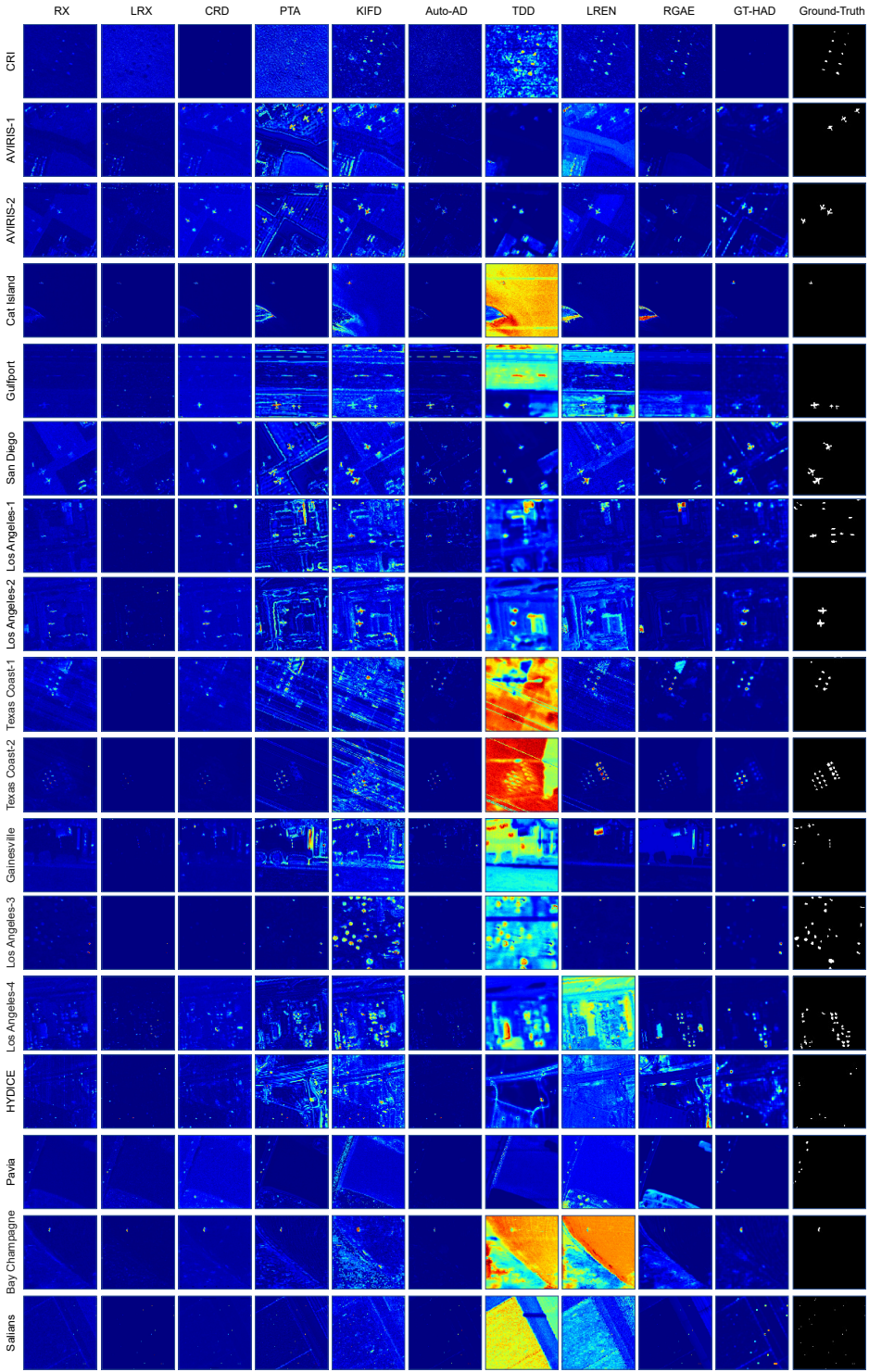} 
    \caption{ Colour Anomaly maps of different HAD methods on 17 datasets}
    \label{fig:heatmap}
\end{figure*}

The performance comparison reveals distinct characteristics and trade-offs among the evaluated HAD algorithms.\\
\textbf{Overall Trends:} Across all datasets, deep learning models, particularly GT-HAD, generally achieve the highest detection accuracy (average AUC: 0.9733). However, this often comes at the cost of higher computational requirements compared to traditional statistical methods. Conversely, the statistical RX algorithm demonstrates exceptional speed (average time: 0.40s), making it highly suitable for real-time and resource-constrained applications, despite its slightly lower average AUC (0.9390).\\
\textbf{Categorical Analysis}
\begin{itemize} {
\item {Statistical methods (RX,LRX): RX consistently leads in computational efficiency. While its AUC is competitive on many datasets such as, CRI, Abu Beach-3, it shows weaker performance on others like Los Angeles-1, Salinas. LRX, despite achieving moderate accuracy (average AUC: 0.9013), often incurs significantly higher computational costs, particularly evident on datasets like CRI (557.93s).}
\item {
Representation-based Models (PTA, CRD): CRD demonstrates strong accuracy, achieving the highest AUC among non-deep learning models (average 0.9567), but it is computationally intensive (average 31.96s). PTA, while slightly faster (average 30.40s), generally shows reduced detection accuracy (average AUC: 0.9141).
}
\item{
Classical Machine Learning (KIFD): KIFD exhibits robust accuracy (average AUC: 0.9529), performing particularly well on complex scenes like HYDICE Urban. However, its practical deployment can be limited by longer inference times (average 57.51s) for achieving  real-time detection.
}
\item {
Deep Learning-based Techniques: GT-HAD consistently emerges as the top performer in terms of accuracy (average AUC: 0.9733), while maintaining a moderate computational time (average 30.51s). Auto-AD offers a balance between accuracy (average AUC: 0.9273) and efficiency (average time: 8.05s), making it a viable choice for many practical tasks. TDD and LREN, while showing competitive performance on some datasets, exhibit highly variable results and, in some cases, TDD on Cat Island, Texas Coast, Pavia; LREN on Abu Beach-3, Abu Urban-5, yield very low AUC scores, suggesting instability or specific dataset limitations. RGAE also shows high computational cost and varied AUC performance.
}}
\end{itemize}
\textbf{Quantitative Analysis:}
Beyond quantitative metrics, visual analyses provide deeper understanding of algorithm behavior.Figure~\ref{fig:heatmap} presents the color-coded anomaly maps generated by each method. Algorithms like GT-HAD and KIFD demonstrate a balanced performance across various datasets such as Salinas, Abu-Urban-4, San Diego, AVIRIS-1, and AVIRIS-2, producing anomaly maps that closely resemble the ground truth, effectively highlighting anomalies while suppressing background clutter.  
\par
\begin{figure*}[htbp!]
    \centering
 \includegraphics[width=0.9\textwidth]{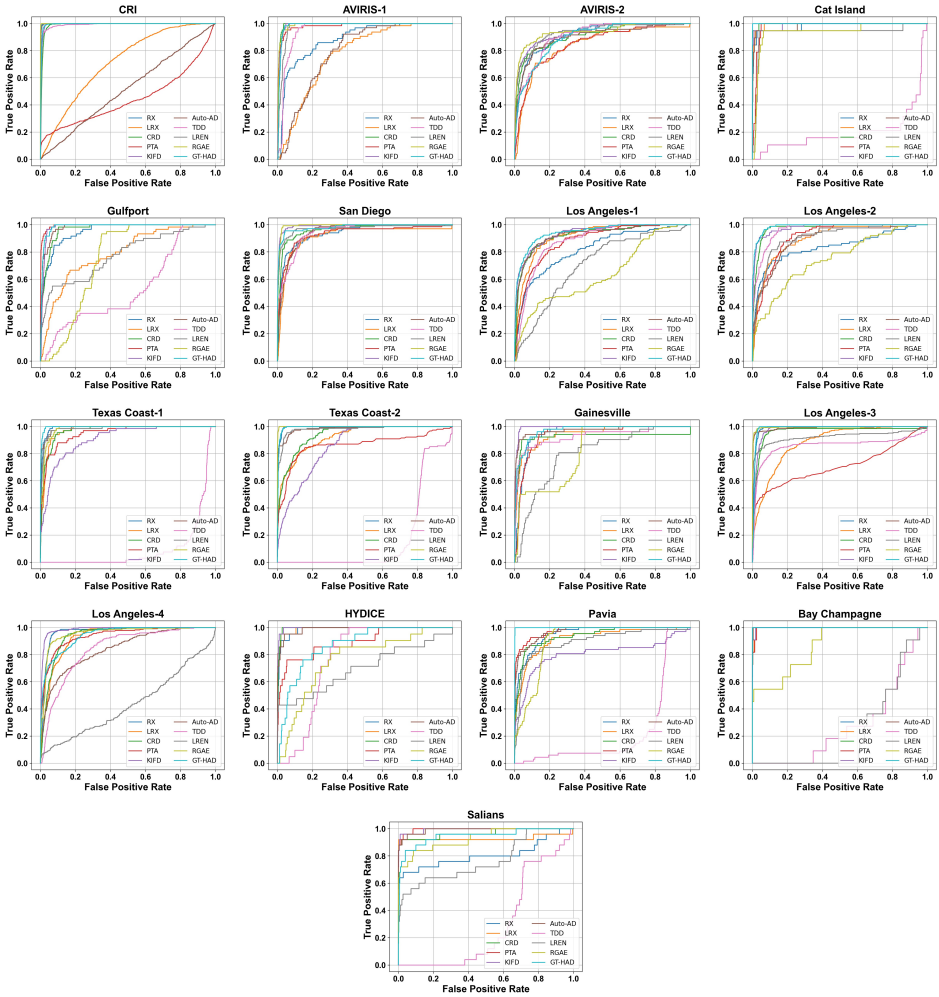} 
    
    \caption{ 2D ROC curves of ten different methods on 17 datasets}
     
    \label{fig:roc}
\end{figure*}

\begin{figure*}[htbp!]
    \centering
    \includegraphics[width=0.9\textwidth]{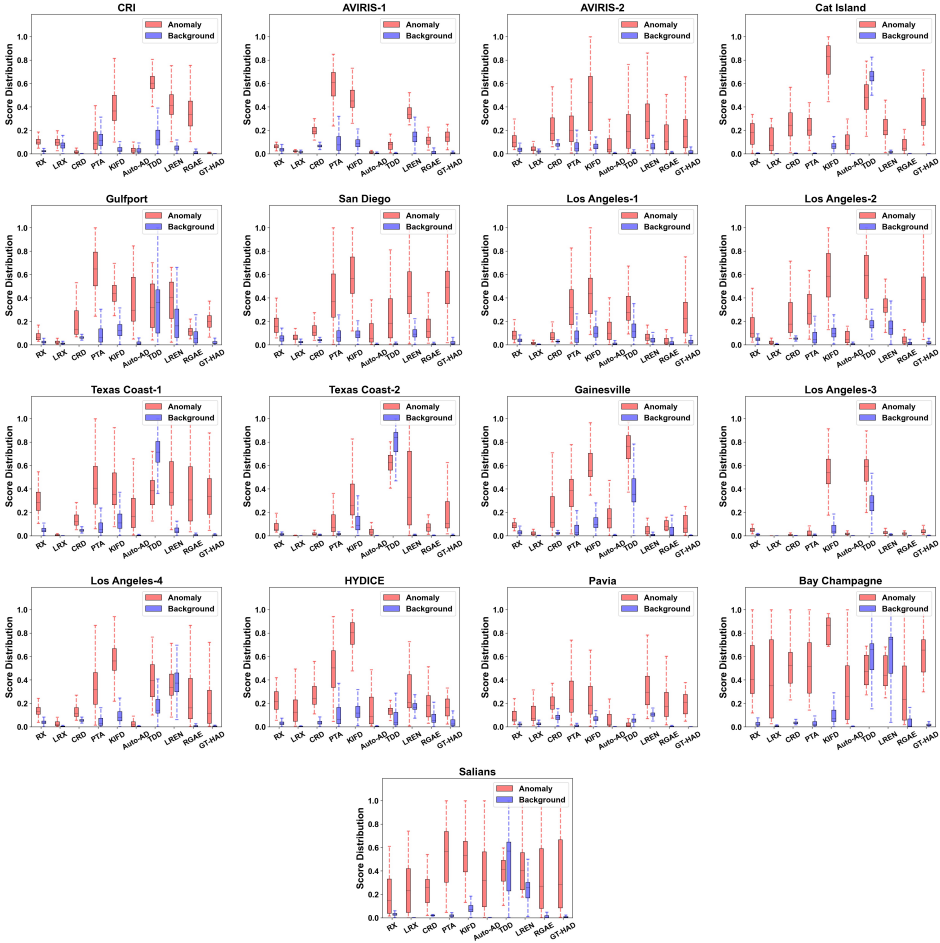} 
     
    \caption{ Box-whisker plots of different HAD methods on 17 datasets}
    
    \label{fig:box}
\end{figure*}

 Similarly, Figure~\ref{fig:box} illustrates the statistical separability maps, where background compression is indicated by narrow box plots (e.g., potentially blue or green in the figure). Methods like KIFD and GT-HAD demonstrate excellent background suppression, with background scores remaining tightly clustered at the lower end across most datasets. This compression enables a clear separation between anomalies and background distributions, minimising false positives. In contrast, methods such as LRX, PTA, LREN, and TDD often produce broader background distributions, which frequently overlap with anomaly scores, leading to poor class separability. This issue is prominently observed in datasets like Gulfport, Abu Urban-5, and Abu Beach-3, where most of these methods struggle to compress the background. 
Furthermore, Figure~\ref{fig:roc} presents the 2D ROC curves, offering a direct visualisation of the trade-off between detection probability and false alarm rate. For instance, on the CRI dataset, while RX shows a near-perfect curve, LRX, Auto-AD, and TDD exhibit notably lower AUC values, indicating poorer overall performance compared to algorithms with high AUCs. Similarly, for the Cat Island dataset, TDD's curve is indicative of the worst performance among the evaluated methods, whereas GT-HAD demonstrates a superior curve, signifying its robust detection capabilities across varying false alarm rates. These visual trends corroborate the quantitative findings presented in Table~\ref{tab:metrics}.

\subsection{Time Complexity Analysis}

In this section, we analyze the time complexity of various algorithms to assess their computational efficiency \cite{Zhou2015FastAD}. Factors such as the number of pixels, image dimensions, filters, and network layers significantly influence a model's runtime. Understanding these factors is crucial for developing real-world solutions that require fast processing. A summary of each algorithm's complexity is provided in Table \ref{tab:time}. According to the table, Classical statistical methods like RX are highly efficient when the number of spectral bands \(K\) is small, making them ideal for deployments. KIFD also shows real-time detection due to its low per-sample complexity and efficient tree-based inference. On the other hand, kernel-based approaches such as KRX and SVDD suffer from high cubic time complexity with respect to the number of samples \(M\), limiting their feasibility in large-scale. Similarly, representation-based techniques like CRD and LRASR, although powerful, are computationally intensive and better suited for offline analysis. Deep learning-based methods, especially those employing autoencoders or lightweight CNNs (Auto-AD, CNND), offer a good trade-off between detection accuracy and runtime performance. Their inference complexity scales linearly with the number of pixels and bands, making them scalable for near-real-time HAD tasks if accelerated by GPUs. The transformer-based GT-HAD, while expensive to train, can be optimised to achieve fast inference, supporting deployment in environmental monitoring systems.

\begin{table*}[htbp!]
\centering
\caption{Time Complexity Analysis of Hyperspectral Anomaly Detection Algorithms}
\label{tab:time}
\footnotesize
\resizebox{\textwidth}{!}{%
\begin{tabular}{lllp{9.5cm}}
\toprule
\textbf{Category} & \textbf{Algorithm} & \textbf{Time Complexity} & \textbf{Remarks} \\
\midrule

\multirow{3}{*}{\textbf{Statistical}}
& RX
& $O(MK^2) + O(K^3)$
& $M$ = number of pixels, $K$ = number of spectral bands. Dominated by \textbf{covariance matrix inversion} and Mahalanobis distance. Efficient when $M \gg K$. \\

& KRX
& $O(M^2K) + O(M^3)$
& Involves \textbf{kernel mapping} on $M$ pixels. Higher complexity due to quadratic or cubic terms in $M$. \\

& LRX
& $O(MnK^2) + O(MK^3)$
& $n$ = number of samples per local window. Repeated local \textbf{covariance estimation} increases cost with large $n$ or $M$. \\

\midrule

\multirow{4}{*}{\textbf{Representation}}
& CRD
& $O(MK^2) + O(M^2K)$
& Collaborative Representation using a dictionary of size $\sim M$. \textbf{Least-squares solution} is the dominant step. \\

& LRASR
& $O(T \cdot \min(MK^2, M^2K))$
& Low-Rank and Sparse Representation. $T$ = number of iterations. \textbf{SVD} or \textbf{ADMM} iterations are computationally intensive. \\

& PTA
& $O(T \cdot MK^2)$
& Prior-based Tensor Approximation. $T$ = decomposition iterations. Primarily driven by \textbf{matrix multiplications}. \\

& RPCA
& $O(T \cdot MK^2)$
& Robust PCA using \textbf{partial SVD}. Generally slower than RX for large $M$ and $K$. \\

\midrule

\multirow{2}{*}{\textbf{Classical ML}}
& KIFD
& $O(MK) + O(M \log M)$
& Kernel Isolation Forest: complexity from feature mapping and \textbf{tree construction}. \\

& SVDD
& $O(M^3)$
& Support Vector Data Description. Solves a \textbf{quadratic programming (QP) problem}, leading to cubic time in $M$. \\

\midrule

\multirow{3}{*}{\textbf{Deep Learning}}
& Auto-AD
& Train: $O(L \cdot MK)$, Inference: $O(L \cdot MK)$
& Autoencoder with $L$ layers. \textbf{Training and inference are linear} in pixel count, making it highly scalable. \\

& CNND
& Train: $O(E \cdot M \cdot F)$, Inference: $O(MF)$
& Convolutional Neural Network. $F$ = number of filters; $E$ = training epochs. \textbf{Requires GPU} for efficient real-time usage. \\

& GT-HAD
& Train: $O(E \cdot M \cdot H)$, Inference: $O(MH)$
& Gated Transformer with $H$ heads/features. \textbf{High training cost} but optimized inference is possible. \\

\bottomrule
\end{tabular}%
}
\end{table*}

\section{Challenges in Hyperspectral Anomaly Detection}~\label{sec:challenges}
Despite significant advancements, HAD still faces several challenges that hinder the development of accurate and efficient detection methodologies. 
A primary issue is the high false positive rate caused by the irregular and rare nature of anomalies, which makes their distinction from complex background elements challenging \cite{Pang_2021}. This problem is compounded by the high-dimensional nature of hyperspectral images where the large number of narrow spectral bands frequently introduces nonlinearity in spectral signatures. This inherent complexity further complicates the process of accurately characterizing and detecting anomalies.

Furthermore, HAD performance is critically affected by external environmental and sensor-induced variations. Factors such as sensor noise, calibration errors, and dynamic environmental conditions (e.g., changes in illumination, atmospheric composition, and temporal variations) cause spectral characteristics to fluctuate \cite{Xu_2022}. Errors in spectral calibration or in the complex process of atmospheric correction (converting radiance spectra to reflectance, or vice versa, which relies on reliable atmospheric data) can introduce significant spectral distortions. Such inaccuracies can misrepresent both anomaly and background spectral signatures, ultimately hindering robust anomaly detection.  Moreover, the spectral appearance of even the same object can vary considerably due to differences in imaging equipment and environmental conditions, further interfering with detection results \cite{Bui_2024}.
In the context of algorithm development, persistent challenges include the creation of real-time solutions that can be easily deployed and are generalised across diverse datasets. While many current anomaly detection algorithms demonstrate strong performance on publicly available datasets, they often have limited access to industry-specific data, which significantly restricts their applicability and deployability in real-world scenarios. Additionally, the high computational time consumption of advanced deep learning models poses a notable hurdle for their practical integration in applications requiring immediate responses. Finally, the limited availability of high-quality, labeled datasets remains a substantial obstacle. This scarcity severely restricts the effective learning of anomaly characteristics, thereby making the development and validation of supervised learning approaches particularly challenging for HAD.

\section{Future Enhancements and Suggestions}~\label{sec:future_enhance}  

Despite significant advancements in HAD, several challenges continue to hinder its widespread adoption and practical deployment in diverse real-world scenarios. Addressing these challenges necessitates focused future research on enhancing real-time capabilities, improving model generalization, integrating robust denoising, and developing automated parameter optimization strategies \cite{1661816}.

\subsection{Need for Real-Time Solutions}  
Most existing HAD techniques face computational costs, due to the large spectral and spatial dimensions inherent in hyperspectral images. This computational burden significantly impedes their applicability in  real-time scenarios, such as remote sensing applications for disaster monitoring and rapid surveillance. Future research should focus optimizing algorithms  using parallel computing, GPU acceleration, and designing more efficient deep learning architectures to enable real-time anomaly detection.  

\subsection{Need for Generalisation Across Different Datasets}  
Many existing anomaly detection models perform well on specific datasets but fail to generalise when applied to different hyperspectral datasets. This lack of generalisation stems from variations in sensor characteristics, diverse environmental conditions, and differing spectral resolutions across datasets.  Future efforts should focus on developing robust domain-adaptive models and advanced transfer learning techniques. Such approaches can significantly improve model robustness, allowing them to generalise more effectively across diverse datasets without substantial performance loss, thereby enhancing their versatility.  

\subsection{Need for Robust Denoising Integrated with Detection}  
Noise present in hyperspectral images can compromise anomaly detection performance. Although various denoising techniques exist, most current models treat denoising and anomaly detection as distinct, sequential tasks. This separation risks removing or distorting crucial anomaly-related information during the noise removal phase. Consequently, future research should concentrate on developing and integrating robust denoising methods directly within anomaly detection frameworks. This integrated approach would ensure that noise suppression actively enhances, rather than degrades, detection accuracy. Exploring hybrid methodologies, such as attention-based filtering and adaptive noise modelling, could further improve detection reliability, especially in challenging noisy environments.

\subsection{Need for Automatic Parameter Optimisation}  
A significant barrier to the usability and scalability of many hyperspectral anomaly detection models is their reliance on extensive manual tuning of hyperparameters, which is often required for optimal performance on different datasets. This manual intervention is time-consuming and expertise-dependent. Therefore, future research should prioritise the development of self-optimising frameworks capable of automatically fine-tuning hyperparameters based on a dataset's unique characteristics. Techniques such as Bayesian optimisation, genetic algorithms, and reinforcement learning-based tuning present promising avenues for achieving automated parameter selection, thereby greatly enhancing model adaptability and reducing user effort.  

\subsection{Other Emerging Research Directions}  

Beyond addressing the aforementioned fundamental needs, several other promising research areas require focused attention to advance HAD. These include: developing lightweight and energy-efficient models, which is crucial for enabling real-time onboard processing capabilities in resource-constrained platforms like Unmanned Aerial Vehicles (UAVs) and satellites. Furthermore, multimodal data fusion offers significant potential by integrating hyperspectral images with complementary data sources such as Light Detection and Ranging (LiDAR), Synthetic Aperture Radar (SAR), or thermal imaging, as this fusion can provide richer contextual information, leading to improved anomaly discrimination. Another vital direction involves exploring self-supervised and few-shot learning techniques to mitigate the strong dependence on extensive labelled anomaly data, a persistent challenge given the inherent rarity of anomalies. Finally, improving visualisation techniques for anomaly detection models is essential, as enhanced visualisation can provide better interpretability of model outputs, fostering greater trust and enabling more informed decision-making by human operators.
By addressing these challenges and pursuing these emerging research directions, hyperspectral anomaly detection can become more efficient, robust, generalizable, and ultimately more impactful for a wide range of real-world applications.

\section{Conclusion}~\label{sec:conclusion}  
In this survey, we conducted a comprehensive review and analysis of diverse techniques for HAD. Our examination spanned a wide spectrum of methodologies, from fundamental preprocessing strategies such as spectral band selection and normalisation, to advanced feature extraction and model architectures. This included statistical methods (global RX, Local RX, Kernel PCA), representation-based techniques (low-rank representation, dictionary learning), classical machine learning models (Isolation Forest, clustering-based), and deep learning-based methods (autoencoders, attention mechanisms, and tensor-based feature extraction). Our analysis highlighted the critical role of noise handling and robust denoising strategies in achieving reliable detection accuracy.
A key insight from our comparative evaluation across 17 benchmark datasets is the trade-off between detection accuracy and computational efficiency. While deep learning models often achieve superior detection performance, classical statistical methods like RX, consistently demonstrate exceptional speed, making them highly suitable for real-time and resource-constrained applications.
Despite these recent advancements, hyperspectral anomaly detection still faces challenges. These include limitations in achieving real-time processing speeds, difficulties in ensuring model generalization across diverse datasets due to inherent sensor and environmental variations, and the persistent need for robust noise mitigation techniques that do not compromise anomaly information.

To propel HAD forward, future research should strategically focus on developing self-adaptive and noise-aware anomaly detection models that can dynamically optimize parameters. Emphasizing multimodal data fusion will provide richer contextual information, while optimizing algorithms for lightweight and real-time deployment on platforms such as UAVs and satellites is crucial for practical implementation. Exploring self-supervised and few-shot learning techniques will also be vital in reducing dependence on scarce labeled data. By addressing these multifaceted challenges, hyperspectral anomaly detection can evolve into a more efficient, robust, generalizable, and scalable technology, unlocking its full potential across a wide array of remote sensing applications.



\section*{Acknowledgments}
We thank the authors of the algorithms used in our empirical comparison for providing their source codes. This work is supported by the Air Force Office of Scientific Research under award number FA2386-23-1-4003. 


\end{document}